\documentclass[11pt,a4paper]{article}
\usepackage{times}
\usepackage{url}
\usepackage{latexsym}
\usepackage[utf8]{inputenc}
\usepackage{graphicx}
\usepackage{xcolor}
\usepackage{amsfonts}
\usepackage[comma]{natbib}
\usepackage{fullpage}
\usepackage{paralist}
\usepackage{subcaption}
\usepackage{booktabs}
\usepackage{multirow}
\usepackage{tipa}
\usepackage{amsmath}
\usepackage{siunitx}

\title{Three tree priors and five datasets: A study of the effect of tree priors in Indo-European phylogenetics}

\author{Anonymous}

 \author{
   Taraka Rama\\
   Department of Informatics\\
   University of Oslo, Norway \\
   {\tt tarakark@ifi.uio.no}
 }

\date{}

\begin{document}

\maketitle

\begin{abstract}
The age of the root of the Indo-European language family has received much attention since the 
application of Bayesian phylogenetic methods by \citet{gray2003language}. The root age of the Indo-European family has tended to decrease 
from an age 
that supported the Anatolian origin hypothesis to an age that supports the Steppe origin hypothesis with the application of new models 
\citep{chang2015ancestry}. 
However, none of the published work in the Indo-European phylogenetics studied 
the effect of tree priors on phylogenetic analyses of the Indo-European family. In this paper, I intend to fill this gap by exploring 
the effect of tree priors on different aspects of the Indo-European family's phylogenetic inference. I apply three tree priors---Uniform, 
Fossilized Birth-Death (FBD), and Coalescent---to five 
publicly available datasets of the Indo-European language 
family. I evaluate the posterior distribution of the trees from the Bayesian analysis using Bayes Factor, and find that there is support 
for 
the Steppe origin hypothesis in the case of two tree priors. I 
report the median and 95\% highest posterior density (HPD)
interval of the root ages for all the three tree priors. A model comparison suggested that either Uniform prior or FBD
 prior is more suitable than the Coalescent prior to the datasets belonging to the Indo-European language family.

%Finally, we offer an explanation for the differences by comparing the equations of the three different priors. 
%Finally, the results in this paper support a date for Indo-European language of about 6500 years old (consistent with 
%the Steppe origin hypothesis of the Indo-European languages).
%The inferred root date of the family shifted from an age that supports 
%the Anatolian origin hypothesis to an age that supports the Steppe origin due to the development of new methods and 
%better data.
\end{abstract}

\section{Introduction}
The Indo-European language family is widely spoken and consists of languages belonging to subgroups such as Albanian, 
Armenian, 
Balto-Slavic, Germanic, Greek, Indo-Iranian, and Italo-Celtic. The root age of the Indo-European family has been a heavily debated topic 
since the application of Bayesian 
phylogenetic methods to lexical cognate data. The root age of the Indo-European language family was estimated using phylogenetic methods 
developed in computational biology 
\citep{gray2003language,atkinson2005words,nicholls2008dated,ryder2011missing,bouckaert2012mapping}. These phylogenetic methods employ 
lexical cognate data 
(from Swadesh word lists [table \ref{tab:cog2bin}]; \citealt{swadesh1952lexico}) and external evidence (from archeology and history) 
regarding both the age of the ancient languages (such as Latin) and the age of the internal subgroups (such as Germanic) to infer the 
timescale of the Indo-European phylogeny. The work of Gray and 
colleagues 
produced root age estimates that supported the Anatolian origin hypothesis (8000--9500 Years Before Present [B.P];  
\citealp{renfrew1990archaeology}) of the Indo-European language family. In contrast, historical 
linguistics---based on cultural and material vocabulary---points to a Steppe origin of the Indo-European language 
family where the root age falls within the range 5500--6500 Years B.P \citep{anthony2015indo}.
 \footnote{The scripts, the data files, and the results of the paper are available at \url{https://github.com/PhyloStar/ie-phylo-exps}.}

In a followup work, \citet{chang2015ancestry} corrected the IELex dataset \citep{dunn2012indo}---originally compiled by 
\citet{dyen1992indoeuropean}---and tested a wide range of 
models and datasets. \citet{chang2015ancestry} modified the Bayesian phylogenetic inference software BEAST 
\citep{drummond2012bayesian} such that the software samples trees that show eight ancient languages---Vedic Sanskrit, 
Ancient Greek, Latin, Classical Armenian, Old Irish, Old English, Old High German, and Old West Norse---as ancestors of modern descendant 
languages (table \ref{tab:ancient}). The results of their analysis showed that the 
estimated median root age of the Indo-European language family falls within the age range that supports the Steppe origin of 
the Indo-European language family. 

\begin{table*}[!ht]
\centering
  \begin{tabular}{lp{10cm}}
\toprule
Ancient language & Modern descendants\\\midrule
 Vedic Sanskrit & Indo-Aryan languages \\
Ancient Greek & Modern Greek \\
Latin & Romance languages \\
Classical Armenian & Modern Armenian dialects: Adapazar, Eastern Armenian \\
Old Irish & Irish, Scots Gaelic \\
Old English & English \\
Old West Norse & Faroese, Icelandic, Norwegian \\
Old High German & German, Swiss German, Luxembourgish \\\bottomrule
\end{tabular}
\caption{Ancestry constraints: ancient languages and their descendants employed 
by \citet{chang2015ancestry}.}
\label{tab:ancient}
\end{table*}

%While discussing their results, \citet{chang2015ancestry} hypothesized that the coalescent tree prior without ancestry 
%constraints infers higher root age than 
%made the following comment about the reason why the ancestry constraints infer a lower root age than a model without 
%ancestry constraints:

%\begin{quote}
%A realistic tree prior must assign a nonzero probability to the set of trees where one language is ancestral to 
%another. (If an ancient language is newly discovered, it might be directly ancestral to a known language.) However, 
%most tree priors in common use [\ldots] assign an infinitesimal probability to that set of trees. Due to our Bayesian 
%methodology, even if the maximum likelihood tree were a tree in that set, we would not find it, since the trees sampled 
%from the posterior distribution would almost surely not be from that set.
%\end{quote}

The phylogenetic dating analyses reported by \citet{bouckaert2012mapping} and \citet{chang2015ancestry} are based on a coalescent tree 
prior that employs both 
the ages of the ancient languages and the internal node ages to infer the dates of all the internal nodes (and the root) of a 
language tree. The coalescent tree prior described in the context of Bayesian phylogenetic inference by 
\citet[309--320]{yang2014molecular} is based on the coalescence process studied by 
\citet{kingman1982coalescent}, and is used to model the spread of viruses or alleles in a population of 
individuals
across time. 

The coalescent tree prior cannot model the linguistic reality that an ancient language such as Old English is the 
ancestor of Modern English. It will infer that both Old English and Modern English descended from an 
unattested linguistic common ancestor. This observation is the departure for the ancestry constrained analyses reported 
by \citet{chang2015ancestry}. The authors found that constraining an ancient language to be the ancestor 
of modern language(s) infers a reduced age for the root of the Indo-European language family which supports the Steppe 
origin hypothesis. 

%\citeauthor{chang2015ancestry} also refer to a paper by \citet{stadler2010sampling}, who proposed a tree prior where an ancient language 
% can either be an internal node or a leaf (tip) in 
%the tree. This tree 
%prior is known as Fossilized Birth-Death (FBD) prior \citep{heath2014fossilized,gavryushkina2014bayesian,zhang2015total} and 
%is an extension to the birth-death prior used in Bayesian phylogenetics \citep{yang1997bayesian,hohna2011inferring}. 

% The dating analysis of \citet{chang2015ancestry} is known as \emph{tip dating} where fossils are also tips of the tree whose age and 
% lexical cognate information is 
% used to infer and calibrate phylogenies. The analysis of \citet{bouckaert2012mapping} uses both internal node age constraints and tip 
% dates to infer the internal node ages of the Indo-European 
% language family. In the context of tip dating, there are two other tree priors, namely, Fossilized Birth-Death (FBD) prior 
% \citep{stadler2010sampling,heath2014fossilized,gavryushkina2014bayesian,zhang2015total} and uniform tree prior \citep{ronquist2012total} 
% that are implemented in MrBayes 3.6 \citep{ronquist2012mrbayes}.

While discussing their results, \citet{chang2015ancestry} observed that the coalescent tree prior without ancestry constraints does not 
sample trees where an ancient language can be the ancestor of modern language(s). Therefore, the coalescent tree prior might not be 
appropriate for modeling the evolution of the Indo-European family. 
This observation marks the departure point of 
the 
analyses reported in this paper where I explore the effect 
of tree priors in the Indo-European phylogenetics. All the previous phylogenetic studies involving the Indo-European family compare the fit 
and 
effect of the age 
of different substitution models such as Covarion, Stochastic Dollo, and a binary state Generalized Time Reversible model. However, none of 
the above studies studies the effect of tree priors on dating of the Indo-European language family.

Therefore, in this paper, I attempt to fill this gap by analyzing all the five publicly available datasets (section \ref{ssec:data}) 
using FBD tree prior, uniform prior, and constant population 
size coalescent prior. 
I perform a Bayes Factor analysis similar to \citet{chang2015ancestry} in section \ref{sec:steppe} and find that the trees inferred with 
FBD prior 
\citep{stadler2010sampling,heath2014fossilized,gavryushkina2014bayesian,zhang2015total} and uniform tree prior \citep{ronquist2012total} 
support the Steppe origin hypothesis of the Indo-European languages. Finally, the root's median age and 95\% highest posterior 
density ages inferred from the coalescent analysis support an Anatolian origin of the Indo-European languages.

Unlike \citet{bouckaert2012mapping} and \citet{chang2015ancestry}, I do not supply the subgroup constraint information 
to the phylogenetic program beforehand, but allow the phylogenetic program to infer the tree topology along with the 
divergence times of the internal nodes. I find that the Bayesian phylogenetic program infers known subgroups \emph{correctly} across tree 
priors. My experiments with FBD and uniform priors show 
that 
ancestry constraints are \emph{not necessary} to infer support for the Steppe origin of the Indo-European family. I also 
performed a model comparison based on the Akaike Information Criterion through MCMC (AICM; \citealp{baele2012improving}) and found that 
both 
uniform and FBD priors fit better than coalescent tree 
prior.

The rest of the paper is organized as follows. I will motivate the appropriateness of FBD prior for the Indo-European family 
diversification 
scenario and describe other tree priors in section \ref{sec:treepr}. I will discuss 
the datasets, substitution model, tree prior settings, Monte Carlo Markov Chain settings, and calculation of Bayes 
Factor support for the Steppe origin hypothesis vs. the Anatolian origin hypothesis in section \ref{sec:methods}. I will present the 
inferred median ages and 95\% highest 
posterior density (HPD) age intervals, Bayes Factors, relevance of ancestry constraints, and 
quality of inferred trees in section \ref{sec:results}. Finally, I will 
conclude the paper in section \ref{sec:concl}.

\section{Tree priors}\label{sec:treepr}

In this section, I will describe the three different tree priors used in the paper. First, I describe the coalescent tree prior in section 
\ref{ssec:coalp}. Next, I will motivate why FBD tree prior is more suitable than the Coalescent tree prior for the Indo-European family in 
section \ref{ssec:bdp}. Finally, 
I describe the uniform tree prior in section \ref{ssec:unifp}. 

\subsection{Constant size coalescent prior}\label{ssec:coalp}

The constant population size coalescent tree prior is dependent on the $\theta~(= 2Pc)$ parameter where $P$ is the 
effective population size and $c$ is the base clock rate. The probability of a tree under this model is 
$\prod_{j=2}^{n}\frac{2}{\theta} \exp(-\frac{j(j-1)}{\theta}t_j)$, where $t_j$ is the time during which there are $j$ 
lineages ancestral to the sequences in the data. Both $P$ and $c$ are sampled in this paper. I note that the constant size population 
prior was also used by \citet[A6,220]{chang2015ancestry} to perform an 
ancestry-constrained phylogenetic analysis which supports the Steppe origin hypothesis.\footnote{I discovered a bug 
in the MrBayes implementation with the coalescent prior that was calculating the Metropolis-Hastings ratio incorrectly. My implementation 
is 
already made available here: 
\url{https://github.com/PhyloStar/mrbayes-coal}.} To 
the best of my knowledge, 
I am not aware of any previous interpretation of coalescent process in a linguistic scenario. I make the following interpretation when 
applying the constant size coalescent prior to 
languages.\footnote{This interpretation is due to Igor Yanovich.} 
According 
to this interpretation, the observed languages are lineages from a large haploid population of individual 
languages where each language is spoken in a community.

\subsection{Birth-Death priors}\label{ssec:bdp}
\begin{figure}[!h]
\centering
%  \includegraphics[scale=0.4]{FBD}
%  \caption{The red dots show fossils and the blue dots show the extant and attested species 
% \citep{zhang2016molecular}.}

\begin{subfigure}{0.4\textwidth}
 \includegraphics[scale=0.5]{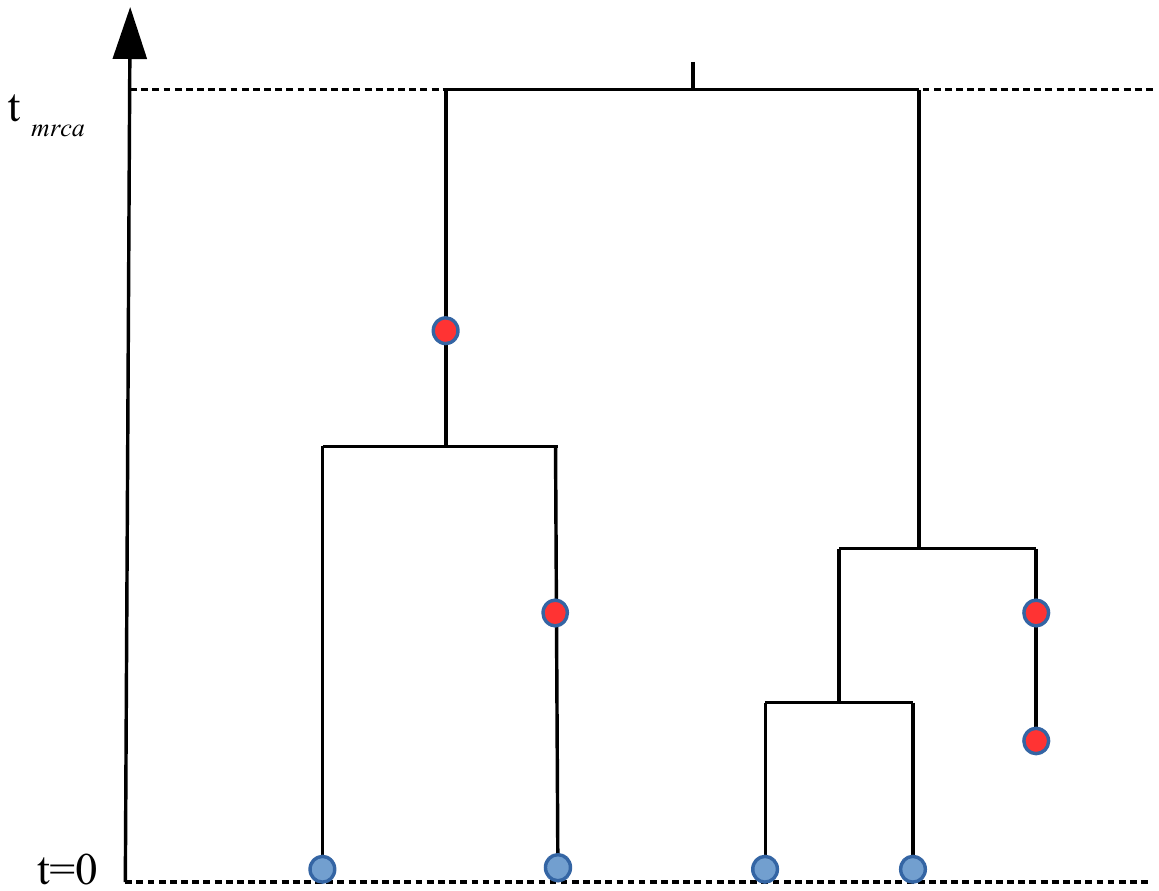}
 \caption{Fossilized Birth-Death tree}
 \label{fig:fbd}
\end{subfigure}
~
\begin{subfigure}{0.4\textwidth}
 \includegraphics[scale=0.5]{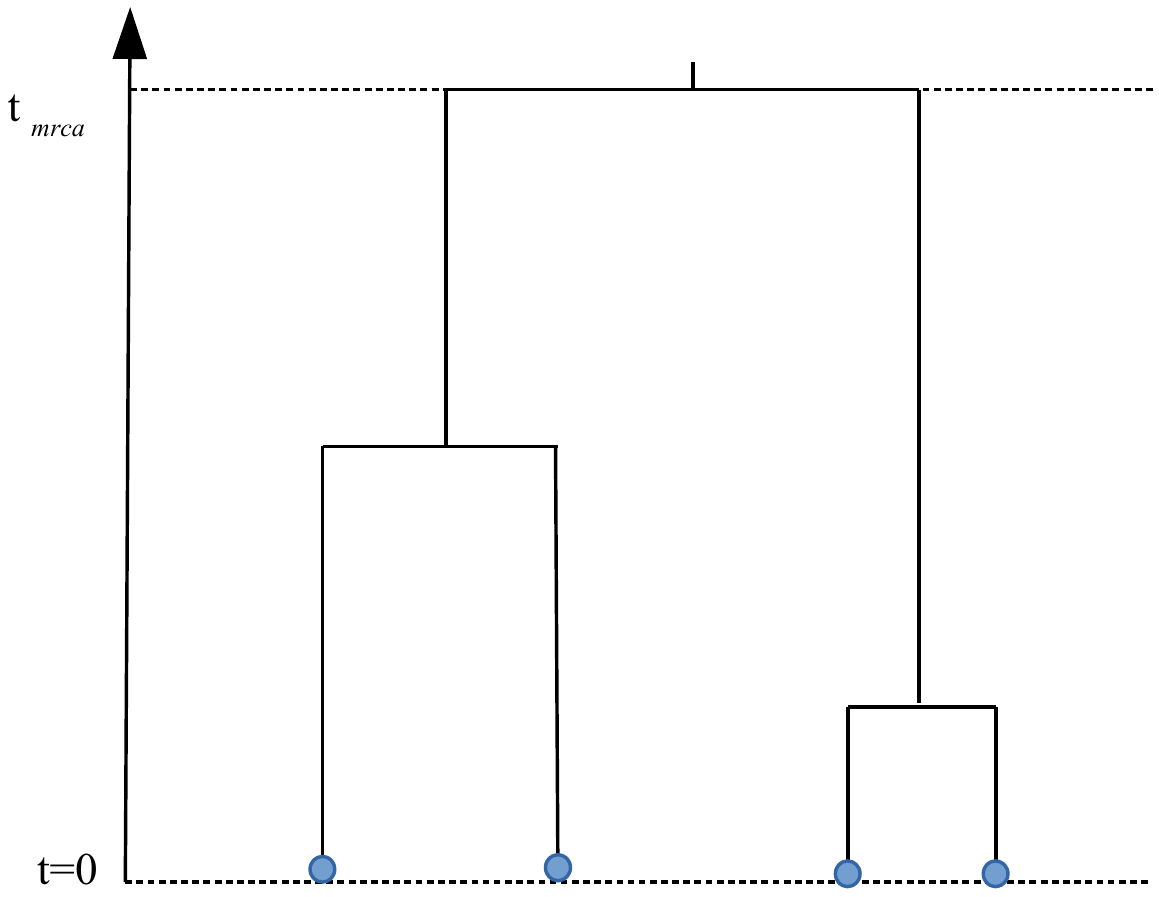}
 \caption{Birth-Death tree}
 \label{fig:bd}
\end{subfigure}

\caption{The red dots show fossils and the blue dots show the extant languages 
\citep{zhang2015total}. The left figure shows the FBD tree 
with fossils as both tips and ancestors of modern languages. The right figure shows the corresponding standard birth-death tree 
with extant languages. $t=0$ shows the present time whereas, $t_{mrca}$ shows the 
age of the most recent common ancestor.}

\label{fig:FBD}
\end{figure}

Birth-Death tree priors are used to model lineage diversification and to date the split event within a phylogeny.
The standard birth-death prior of \citet{yang1997bayesian} is conditioned on the age of the most recent common ancestor ($t_{mrca}$) and 
assumes that 
birth ($\lambda$) and death ($\mu$) rates are 
constant over time. In this model, all the tips in the tree are extant and do not contain any fossils (figure \ref{fig:bd}). A fossil 
can be the ancestor of a modern 
language or can be extinct without leaving any descendants. For instance, Vedic 
is considered to be the ancestor of all the modern Indo-Aryan languages (table 
\ref{tab:ancient}), whereas, Hittite or Gothic are languages that died out 
without leaving any descendant. 

The 
birth-death model described by \citet{yang1997bayesian} handles incomplete languages sampling through $\rho = 
\frac{n}{N}$ where $n$ is the number of languages in the sample and $N$ is the total number of extant languages in the 
family. The birth-death model estimates the species divergence times on a relative scale. The relative times can be 
converted into geological time scale by tying one or more internal nodes to known historical or archaeological evidence. It has to be noted 
that the coalescent process is 
mathematically different from birth-death process \citep[62--63]{stadler2009incomplete}.

In the case of the Indo-European language family, the standard birth-death tree prior of \citet{yang1997bayesian} \emph{only} uses the 
internal 
node calibrations (for instance, the information that Germanic subgroup is about 2200 years old 
\citep{chang2015ancestry}) to infer the remaining internal nodes' dates. This procedure is known as \emph{node dating} 
and has been used for inferring the phylogeny of Bantu languages\footnote{To be precise, 
the scholars used a pure birth (Yule) process with $\mu =0, \rho=1$, a special case of birth-death process, to estimate the 
divergence times of the internal node splits in the Bantu language family phylogeny.} \citep{grollemund2015bantu} and Turkic languages 
\citep{hruschka2015detecting}.\footnote{\citet{hruschka2015detecting} use cognate sets from etymological  dictionary 
where the reflexes within a cognate set need not have the same meaning. This approach is different from the phylogenetic approaches 
used in this and other papers, where the cognates are root-meaning pairs derived from Swadesh lists \citep[201]{chang2015ancestry}.}

The 
node dating 
method does not utilize the available 
lexical cognate information about attested ancient languages that went extinct (e.g. Gothic) or evolved into modern 
languages (e.g. Latin). However, the node dating method indirectly uses the age information of extinct languages to apply constraints to 
the internal node ages of a language family. In another argument against node dating, \citet{ronquist2012total} noted that if there is 
more than one fossil in the same language group, then, only the oldest fossil provides the age constraint for the associated internal node. 
For example, in the case of the Germanic subgroup, there are four fossil languages---Gothic, Old High German, Old English, and Old 
West Norse---out of which only Gothic's age information would be used to specify the minimum age of the Germanic subgroup, whereas the rest 
of 
the fossil languages cannot provide extra information 
regarding the age of the Germanic subgroup.

\citet{stadler2010sampling} proposed an extension to the standard birth-death prior that can handle the placement of 
ancient languages as tips or as internal nodes (fossils; figure \ref{fig:fbd}). This prior is known as Fossilized Birth-Death (FBD) 
Prior since it can handle both fossil and extant species in a single model. The FBD family of priors can model the 
linguistic fact that Old English is the ancestor of Modern English. Along with the parameters, $\lambda$ and $\mu$, the FBD prior also 
features fossil sampling rate parameter $\psi$, which is the rate at which fossils are observed along a 
branch.
% \footnote{\citet{gavryushkina2014bayesian} extended the fossilized-birth-death prior to allow birth, death, and 
% sampling rates to change over time.} 
The FBD tree prior requires only the ages of fossils to infer the root age of a tree; and, is more objective than node dating that requires 
internal node age constraints that are not directly observed. The standard birth-death prior conditioned on $t_{mrca}$ is a special case of 
FBD prior when $\psi=0$ \citep[401]{stadler2010sampling}. An example of a fossilized birth-death tree 
is presented in figure 
\ref{fig:fbd}.
The left 
tree (\ref{fig:fbd}) in figure \ref{fig:FBD} shows the FBD tree including lineages with sampled extant and fossil languages whereas the 
right figure shows the standard birth-death tree with extant languages.

The probability of a tree under the FBD tree prior is conditioned on $t_{mrca}$ and the nature of extant taxa sampling. In this paper, I 
assume that the 
extant taxa are sampled uniformly at random. Unlike \citeauthor{chang2015ancestry}, who impose ancestry constraints 
externally, the FBD tree 
prior can infer the ancestry constraints from the data (if such a signal exists) and do not have 
to be supplied beforehand. The species sampling probability $\rho$ is determined as the ratio 
between the number of extant languages in the dataset to the total number of extant Indo-European 
languages. 

The probability of the tree under the FBD model \citep[equation 5]{stadler2010sampling} conditioned on $x_1$ ($t_{mrca}$) is given 
below. Here, $n~(>1)$ is the 
number of extant sampled tips, $m~(\ge0)$ is the number of extinct sampled tips, $k~(\ge0)$ is the number of sampled ancestors with sampled 
descendants, and $y_i$ is the age of a extinct sampled tip. 

\begin{equation}
 \frac{\lambda^{n+m-2}\psi^{k+m}}{(1-\hat{p}_0(x_1))^2} p_1(x_1) \prod_{i=1}^{n+m-1}p_1(x_i) \prod_{i=1}^{m}\frac{p_0(y_i)}{p_1(y_i)}
\end{equation}

Here, $p_0(t)$, $p_1(t)$, $c_1$, $c_2$, and $\hat{p}_0(x_1)$ are defined as followed:
\begin{itemize}
 \item $p_0(t)$ is the probability that an individual present at time $t$ before present has no sampled extinct or extant descendants, 
which 
is given as \\$\displaystyle p_0(t) = \frac{\lambda+\mu+\psi+c_1\frac{(\exp(-c_1t)(1-c_2))-(1+c_2)}{(\exp(-c_1t)(1-c_2))+(1+c_2)}}{2\lambda}$ 
%  \begin{itemize}
%   \item $\displaystyle p_0(t) = \frac{\lambda+\mu+\psi+c_1\frac{(\exp(-c_1t)(1-c_2))-(1+c_2)}{(\exp(-c_1t)(1-c_2))+(1+c_2)}}{2\lambda}$ 
%  \end{itemize}

 \item $p_1(t)$ is the probability that an individual present at time $t$ before present has only one sampled extant descendant and no 
sampled extinct descendant, which is given as \\$\displaystyle p_1(t) = \frac{4\rho}{2(1-c_2^2)+\exp(-c_1t)(1-c_2)^2+\exp(c_1t)(1+c_2)^2}$
% \begin{itemize}
%  \item  $p_1(t) = \frac{4\rho}{2(1-c_2^2)+\exp(-c_1t)(1-c_2)^2+\exp(c_1t)(1+c_2)^2}$
% \end{itemize}
\item  $\hat{p}_0(x_1) = p_0(t|\psi=0)$, $c_1 = |\sqrt{(\lambda-\mu-\psi)^2+4\lambda\psi}|$, $c_2 = 
-\frac{\lambda-\mu-2\lambda\rho-\psi}{c_1}$
\end{itemize}

FBD tree priors have been used for estimating divergence times for datasets with extant and fossil 
species \citep{heath2014fossilized,gavryushkina2014bayesian,zhang2015total}. Since the Indo-European family has both fossils and extant 
languages, the FBD tree prior that handles attested fossil 
ancestors is 
more suitable than the coalescent tree prior that places fossils as tips. For instance, Tocharian languages went extinct without leaving 
any 
modern 
descendant language, whereas modern Romance languages are the 
descendants of Latin (an ancient language). Moreover, the data for the Indo-European language family comes from divergent languages and not 
from a single 
population. These arguments support the choice of FBD prior over a coalescent prior for modeling the evolution of the Indo-European 
language 
family.

\subsection{Uniform tree prior}\label{ssec:unifp}
Similar to the coalescent tree prior, the uniform tree prior \citep{ronquist2012total} places fossils as tips of the 
tree. However, the uniform tree prior does not make any assumptions regarding the lineage diversification process. The uniform tree prior 
assumes that the internal nodes' ages are uniformly distributed between tip ages and the root age. The prior probability of a tree under 
uniform model is conditioned on the root age $r$ which is 
drawn from a prior distribution $h$. 
Under this model, an interior node age is drawn from a uniform distribution with a tip age as the lower bound and the root age as the 
upper bound. The probability of a tree under the uniform model is proportional to $h(r) \prod_{j=1}^{n-2}\frac{1}{r-t_{j+1}}$ where $t_j$ 
is the age of a tip $j$.
% The uniform 
% prior does not have any parameters and does not make any assumptions regarding the biological process that generates the tree.

\section{Methods}\label{sec:methods}
In this section, I describe the datasets, prior settings, inference procedure details, and calculation of the Bayes Factor.

\subsection{Data}\label{ssec:data}
\begin{table*}[ht]
\centering
% \small
 \begin{tabular}{lc|lc}
\toprule
Language & Age Prior & Language &  Age Prior \\\midrule
 Hittite  & $3500-3600$ & Old High German\textsuperscript{A} & $1000-1100$\\
 Old Irish\textsuperscript{A} & $1100-1300$ & Tocharian B & $1200-1500$\\
 Classical Armenian\textsuperscript{A} & $1300-1600$ & Tocharian A & 
$1200-1500$ 
\\
 Ancient Greek\textsuperscript{A} & $2400-2500$ & Lycian & $2350-2450$\\
  Luvian & $3275-3425$ & Old Prussian & $500-600$\\
 Vedic Sanskrit\textsuperscript{A} & $3000-3500$ & Umbrian & $2100-2300$\\
 Old English\textsuperscript{A} & $950-1050$ & Avestan & $2450-2550$\\
 Old Persian & $2375-2525$ & Gothic & $1625-1675$\\
 Latin\textsuperscript{A} & $2100-2200$ & Old Norse\textsuperscript{A} & 
$750-850$\\
 Oscan & $2100-2300$ & Old Church Slavonic & $950-1050$\\
Cornish & $300-400$ & Sogdian & $1200-1400$\\
\bottomrule
 \end{tabular}
\caption{\label{tab:fossil}Calibration dates for the ancient/medieval languages. All 
dates are given as years before present (BP). The superscript \textsuperscript{A} denotes those 
languages that are assumed to be ancestors of extant languages by 
\citet{chang2015ancestry}.}
\end{table*}

All the five datasets used in this paper---B1, 
B2, \textsc{Broad}, \textsc{Medium}, and \textsc{Narrow}---are assembled from IELex by \citet{chang2015ancestry}.\footnote{One of the 
reviewers asked why I did not experiment with \texttt{CoBL} database
(\url{http://www.shh.mpg.de/207610/cobldatabase}). The database is not publicly 
available to perform experiments.} The B1 dataset is derived from \citet{bouckaert2012mapping} and consists of 207 
meanings for 103 languages. The B2 dataset consists of 97 languages and is a subset of the B1 dataset. The B2 dataset 
is obtained after discarding six languages (Lycian, Oscan, Umbrian, Old Persian, Luvian, and Kurdish) 
that have attestation in less than 50\% of the meanings.

The \textsc{Broad} dataset consists of 94 languages and 197 
meaning classes. The \textsc{Broad} dataset is corrected for cognate judgments in the Indo-Iranian 
subgroup; and, also has an extra medieval language, Sogdian, 
which is not present in B1. Ten meanings that are susceptible to sound symbolism and have poor 
coverage in terms of number of languages are also removed from the 
\textsc{Broad} dataset \citep[213]{chang2015ancestry}. The 
\textsc{Medium} dataset is a subset of the \textsc{Broad} dataset and is assembled in such a way that 
the languages and meanings with poor coverage are excluded. The \textsc{Medium} dataset has 82 
languages 
and 143 meanings. The \textsc{Narrow} dataset is a subset of the \textsc{Medium} dataset and consists of 
only those modern languages that have an attested ancestor. This selection leaves the \textsc{Narrow} dataset 
with 52 languages.\footnote{All the datasets are available at 
\url{http://muse.jhu.edu/article/576999/file/supp02.zip}.}

\subsection{Substitution models}
Bayesian phylogenetics originated in evolutionary biology and works by inferring the evolutionary relationship (trees) 
between DNA sequences of species. The same method can also be applied to binary (morphological) traits of species 
\citep{yang2014molecular}. Linguistic data is binary trait data where each column in the trait matrix is a cognate class. Words that belong 
to 
the same cognate class are coded as \texttt{1}, else, they are coded as \texttt{0}.  For example, in the case of German, French, Swedish, 
and 
Spanish, the word for \emph{all} in German [al\textipa{@}]
and Swedish [\textipa{"}al\textipa{:}a] would belong to the same cognate set as English, 
while French [tu] and Spanish [to\textipa{D}o] belong to a different cognate set. The binary trait matrix for these languages for the 
meaning \emph{all} is shown in table~\ref{tab:cog2bin}. If a language is missing in a cognate set, then the entry for that language is 
coded 
as 
\texttt{?}, and is ignored in the calculation of likelihood using pruning algorithm \citep[255]{felsenstein2004inferring}. I used a 
Generalized Time Reversible model (equivalent to a F81 
model in the case of binary traits) with ascertainment bias correction 
\citep{felsenstein1992phylogenies,lewis2001likelihood} for all unobserved \textbf{0} columns. The rate variation across sites is 
modeled using a discrete Gamma model with four rate categories \citep{yang1994estimating}, where the shape parameter of the Gamma 
distribution is 
drawn from a exponential prior with mean $1$.

\begin{table}[!ht]
 \centering
    \begin{subtable}{.4\linewidth}%{lccc}
     \centering
    \begin{tabular}{lccc}
   \toprule
      Language & ALL & AND & $\ldots$\\
          \midrule
    English & \textipa{O:}l\textsuperscript{1} & \textipa{ae}nd\textsuperscript{1}  & $\ldots$\\
    German & al\textipa{@}\textsuperscript{1} & \textipa{U}nt\textsuperscript{1} & $\ldots$\\
    French & tu\textsuperscript{2} & e\textsuperscript{2}   & $\ldots$\\
    Spanish & to\textipa{D}o\textsuperscript{2} & i\textsuperscript{2}  & $\ldots$\\
    Swedish & \textipa{"}al\textipa{:}a\textsuperscript{1} & \textipa{O}k\textipa{:}\textsuperscript{3} & $\ldots$\\
       \bottomrule
    \end{tabular}
    \caption{Forms and cognate classes}\label{tab:cogclas}
    \end{subtable}
    \quad
    \begin{subtable}{.4\linewidth}%{lccccc}
    \centering
    \begin{tabular}{lccccc}
   \toprule
      Language & \multicolumn{2}{c}{ALL} & \multicolumn{3}{c}{AND}\\
          \cmidrule(lr){1-1}\cmidrule(lr){2-3}\cmidrule(lr){4-6}
    English & 1 & 0 & 1 & 0 & 0 \\
    German & 1 & 0 & 1 & 0 & 0 \\
    French & 0 & 1 & 0 & 1 & 0 \\
    Spanish & 0 & 1 & 0 & 1 & 0 \\
    Swedish & 1 & 0 & 0 & 0 & 1 \\
       \bottomrule
     \end{tabular}
    \caption{Binary Matrix}\label{tab:binmat}
    \end{subtable}
\caption{Excerpt from meaning list showing cognate classes (table \ref{tab:cogclas}) and the binary cognate matrix (table \ref{tab:binmat}) 
for meanings ALL and AND in five languages. The superscript indicates words that are cognate.}
\label{tab:cog2bin}
\end{table}

\subsection{Tree prior settings}
In this paper, I assumed that the extant languages are randomly sampled. The FBD tree prior is dependent on the number of extant
languages in the sample. I estimated the number of extant Indo-European languages (400) from Glottolog 
\citep{nordhoff2011glottolog}, and set the $\rho$ parameter accordingly for each 
dataset. For FBD prior, the net diversification rate $d ~(=\lambda-\mu)$ is drawn from a exponential prior with mean $1$, 
the relative extinction rate (turnover) $r~(=\mu/\lambda)$ is drawn from a Beta(1,1) prior, and the fossil sampling 
probability $f~(=\psi/(\psi+\mu))$ is also drawn from a Beta(1,1) prior.

I draw the root age from a uniform distribution bounded between $4000$ and $25000$ years in the case of FBD and uniform priors. The root 
age's upper bound is fixed at $25000$ years since this age is 
more than double the upper
bound of the age limit of the Anatolian origin hypothesis. In fact, none of the inferred trees' root ages are close to $25000$ years. The 
coalescent prior, as implemented in MrBayes, is not 
conditioned on $t_{mrca}$. All the fossils' age priors were drawn from uniform distributions whose age ranges are given in table 
\ref{tab:fossil}. 

In the case of the coalescent prior, population parameter $P$ is drawn from a Gamma distribution with shape parameter $1$ and rate 
parameter 
$0.01$. The base clock rate $c$ is drawn from an exponential prior with mean $10^{-4}$. In all the analyses, I use a Independent 
Gamma Rate model \citep{lepage2007general}, where each branch rate is drawn from a Gamma distribution with mean $1.0$ and variance 
$\sigma^2_{IG}/b_j$, where $b_j$---the branch length of a branch $j$---is computed as the product of geological (or calendar) time 
$t_j$ and $c$. 
$\sigma^2_{IG}$ 
is the independent gamma rate model's variance parameter that is drawn from an exponential prior with mean $0.005$. I do not 
employ topology constraints and allow the software to infer the Indo-European phylogeny along with the time scale from the 
data.

\subsection{Markov chain Monte Carlo sampling}
I ran all the experiments using MrBayes software.\footnote{Available at \url{http://mrbayes.sourceforge.net/}.} I ran two independent runs 
(each run consisted of one cold chain and two hot chains) and 
verified that the average 
standard deviation of split frequencies \citep{ronquist2012mrbayes} between both the runs is less than $0.01$. I ran 
all the analyses for 20--80 million 
states and sampled every $1000^{th}$ state to reduce auto-correlation between the 
sampled states. For each dataset, I threw away the initial 25\% of the states as burn-in and generated a 50\% majority 
rule consensus tree\footnote{A 50\% majority consensus tree is a summary tree that consists of only those clades that 
occur in more than 50\% of the post burn-in sample of trees.} from the remaining 75\% of the 
states \citep[chapter 30]{felsenstein2004inferring}.\footnote{I also present the inferred 
phylogenies, posterior support and HPD intervals of the internal nodels for all the tree priors and datasets in the 
appendix.}
% All the code to calculate Bayes Factor, MrBayes scripts, data, and the consensus trees 
% are available at \url{https://doi.org/10.5281/zenodo.829741}}

% Write about the languages that are excluded from Chang et al.\\

\subsection{Evaluating Steppe vs. Anatolian Hypothesis}\label{sec:steppe}
For each dataset, I ran the MrBayes software twice: once without cognate data to generate a prior 
sample of trees and once with cognate data to generate a posterior sample of trees. Then, I used 
Bayes Factor (BF) formulation from \citet{chang2015ancestry} to calculate the support for 
respectively the Anatolian (A) and Steppe (S) hypothesis. Given data $D$, the Bayes factor 
$K_{S/A}$ is calculated as follows:

\begin{equation}\label{eq:ksa} 
\frac{\mathbb{P}(D|t_R \in \Omega_{S})}{\mathbb{P}(D|t_R \in 
\Omega_{A})}
\end{equation}

where, $\Omega_{S} \in [5500, 6500]$ and $\Omega_{A} \in [8000, 9500]$ represents the range of 
Steppe and Anatolian ages and $t_R$ denotes the root age of a tree which is $t_{mrca}$ in the case of FBD prior. The numerator and 
denominator in 
equation \ref{eq:ksa} are computed as follows:

\begin{equation}\label{eq:ksa1}
 \frac{Pr\{t_R \in \Omega_{S}|D\}}{Pr\{t_R \in \Omega_{S}\}}/\frac{Pr\{t_R \in 
\Omega_{A}|D\}}{Pr\{t_R \in \Omega_{A}\}}
\end{equation}

The numerators $Pr\{t_R \in \Omega_{S}|D\},Pr\{t_R \in \Omega_{A}|D\}$ in equation \ref{eq:ksa1} correspond to the fraction of trees in 
the posterior 
sample for which $t_R \in \Omega_S$ and $t_R \in \Omega_A$. The denominators $Pr\{t_R \in \Omega_{S}\}, Pr\{t_R \in \Omega_{A}\}$ 
correspond to the 
fraction of trees in the prior sample for which $t_R \in \Omega_S$ and $t_R \in \Omega_A$. 
Following the interpretation of Bayes Factor by \citet{kass1995bayes}, the 
support for Steppe origin hypothesis is very strong if $K_{S/A} > 150 $, strong if $20 < 
K_{S/A} < 150$, positive if $3 < K_{S/A} < 20$, not worth more than a bare 
mention (\emph{neutral}) if $1 < K_{S/A} < 3$ and negative if $K_{S/A} < 1$.

\section{Results}\label{sec:results}
In this section, I present and discuss the root's median age and 95\% HPD age intervals, fit of tree prior, Bayes Factor support 
for the Steppe vs. the Anatolian hypotheses, comparison of subgroups' inferred dates with expert 
dates, relevance of clade constraints, and ancestry constraints.

\subsection{Median and 95\% HPD ages}
%The HPD interval of the root age for \textsc{narrow} 
%and \textsc{broad} datasets are in the 
%similar range. 
\begin{table}[!ht]
\centering
% \scriptsize
\begin{tabular}{lcccccc}
\toprule
\multirow{2}{*}{Dataset} & \multicolumn{3}{c}{95\% HPD} & \multicolumn{3}{c}{Median 
Age}\\\cmidrule(lr){2-4}\cmidrule(lr){5-7}
 & FBD & Coalescent & Uniform & FBD & Coalescent & Uniform \\\midrule
\textsc{B1} & 6244--8766 & 8370--11695 & 5760--8115 & 7512 & 9821 & 6789 \\
\textsc{B2} & 6150--8430 & 7590--10913 & 5536--7986 & 7177 & 9133 & 6738\\
\textsc{broad} & 5591--7585 & 6654--9327 & 5073--6947 & 6551 & 7984 & 5935\\
\textsc{medium} & 5942--7921 & 7070--9818 & 5395--7392 & 6845 & 8345 & 6339\\
\textsc{narrow} & 5790--7984 & 6826--9791 & 5423--7646 & 6826 & 8228  & 6462 \\
\bottomrule
 \end{tabular}
\caption{\label{tab:HPD} Columns 2--4 show the 95\% Highest Posterior Density (HPD) and columns 5--7 show the median ages (in 
years before present) of the root node from the consensus tree for each dataset and a tree prior.}
\end{table}

Table \ref{tab:HPD} shows the HPD intervals and median root ages for all dataset and tree prior combinations. None of the reported HPD age 
intervals lie completely within the Steppe age interval or the Anatolian age interval. The lower bounds of HPD 
ages in the case of FBD and uniform priors fall within the Steppe interval, whereas the lower bound of the coalescent prior's HPD interval 
falls beyond the Steppe age interval. In the case 
of \textsc{narrow} and \textsc{medium} datasets, the root age is further reduced to 6826 and 6845 years 
respectively in the case of FBD prior. The median ages inferred by FBD prior belong neither to the Steppe hypothesis interval 
nor to the Anatolian hypothesis interval for all the datasets. The median age inferred by uniform prior for \textsc{Broad}, 
\textsc{Medium}, 
and \textsc{Narrow} datasets lie within the range of the Steppe interval. All the priors infer median ages that lie beyond the Steppe 
interval in 
the case 
of B1 and B2 datasets. The coalescent prior infers root ages that lie within the Antolian hypothesis in the case of all 
the datasets except B1 dataset. Across all the priors, the median root ages decrease when the datasets 
are corrected for errors. The descreasing trend in the median ages is similar to the trend observed in \citet{chang2015ancestry}.

\paragraph{Why the \textsc{broad} dataset yields younger ages?} \citet{chang2015ancestry} argue that sparsely attested languages can 
influence the chronology estimates. The authors argue 
by observing that the ascertainment bias correction to the likelihood calculation \citep{felsenstein1992phylogenies} accounts for 
unobserved cognate sets that are not 
observed in the data, but, does not account for the missing entries in a dataset. For example, if 50\% of the data is missing for a 
language, then the ascertainment bias correction does not account 
for missing 50\% of the data. If there are $x$ unique cognate sets in the observed 50\% of the data, then, there is a possibility that 
the unobserved 50\% of the data also has $x$ unique cognate sets that 
do not enter the likelihood calculation. 

The likelihood calculation would only consider the observed $x$ 
unique cognate sets, therefore, underestimating the true number of unique cognate 
sets for a language
in a dataset. Due to this reason, a language with higher number of missing entries is treated as more 
conservative (or lesser number of character changes) than it should be. This is particularly true for languages such as Hittite, Tocharian 
A 
\& B
which have about $11.95\%$ and $32.74\%$ missing entries in the case of the \textsc{broad} dataset as compared to $1.92\%$ and $2.13\%$ in 
the case of the \textsc{medium} dataset. Since, both Hittite and 
Tocharian doculects are very close to the root of the Indo-European tree, this 
underestimation of number of unique cognate sets leads to a shorter branch length which causes the median root age to be younger. Both 
coalescent and FBD tree priors infer a younger age for \textsc{broad} dataset 
than \textsc{medium} and \textsc{narrow} datasets.

\paragraph{Why the B2 dataset yields younger ages?} The B1 dataset features six sparsely attested 
languages---Lycian, Oscan, Umbrian, Old  Persian, Luvian, and Kurdish---where more than 50\% of the meanings are unattested. As 
explained in the previous paragraph, inclusion of 
sparsely attested languages causes the Bayesian 
inference program to underestimate the root age. The opposite happens when a language has more number of unique cognates than it should 
have. This is the case of 
Luvian, where 33\% of the attested cognate sets are erroneously coded as unique cognate sets, although, they are cognate with either 
Hittite 
or Lycian. This erroneous coding causes the Bayesian software to treat Luvian which is one internal node away from the root node to have 
evolved 
more and posits 
longer branches, therefore, pushing 
the root age of the tree away from the Steppe age interval. The B2 dataset excludes the six sparsely attested languages including 
erroneously coded Luvian
which leads to shortening of the median 
root age in the posterior sample. This effect is clearly observed with both the median root age and 95\% HPD age range in the B2 dataset. 
The 
median root age is pushed 400 years downwards towards the Steppe hypothesis in the case where the FBD tree prior is applied to the B2 
dataset. The 
coalescent prior also infers a younger median age for B2 dataset than B1 dataset, whereas the uniform prior is not 
influenced by the six sparsely attested languages.

\subsection{Which tree prior is the best?}

\begin{table}[!ht]
\centering
% \scriptsize
\begin{tabular}{lccccc}
\toprule
Tree Prior & B1 & B2 & \textsc{broad} & \textsc{medium} & \textsc{narrow} \\\midrule
Uniform Prior & \textbf{94002.748} & 90299.551 & \textbf{89269.61} & 50769.888 & \textbf{32162.117} \\
FBD Prior & 94005.297 & \textbf{90297.721} & 89270.359 & \textbf{50764.79} & 32163.007\\
Coalescent Prior & 94117.099 & 90396.491 & 89374.335 & 50917.074 & 32241.019\\
\bottomrule
\end{tabular}
\caption{\label{tab:AICM} AICM values for each of the datasets. The lower the value the better is the model's fit to 
the data. The best fitting model's AICM value is shown in bold and is computed using \textsc{tracer} \citep{rambaut2007tracer}.}
\end{table}

% Although, the Bayes Factor results suggest that both Uniform and FBD priors 
% support Steppe hypothesis, it is not evident if both the priors fit better or worse than the coalescent prior. 
I 
determine the best model through Akaike Information Criterion through MCMC (AICM; \citealp{baele2012improving}). It has to be noted that 
\citet{bouckaert2012mapping} employ 
both Harmonic Mean and AICM to perform model comparison. In this paper, I only use AICM, since, it is more accurate than 
harmonic mean which is unstable. On the other hand, methods such as stepping stone 
sampling \citep{xie2010improving} and thermodynamic integration \citep{lartillot2006philippe} used to estimate marginal likelihood are more 
accurate than AICM but are computationally intensive and require 
at most $K$ times (usually set to 10) the computation as the original MCMC runs \citep[258--259]{yang2014molecular}.

The AICM values for each dataset and tree prior are presented in table \ref{tab:AICM}. The results show that the Uniform tree prior fits 
the 
best for B1, \textsc{Broad}, and \textsc{narrow} 
datasets. The difference between AICM values of Uniform 
and FBD priors is almost negligible in the case of \textsc{Broad} and \textsc{narrow} datasets. The coalescent prior shows the highest AICM 
value and differs by a large margin when compared with FBD and Uniform priors. Since uniform tree prior has fewer parameters than FBD 
prior, I suggest that any future phylogenetic experiment \emph{should test 
uniform tree prior as a baseline} before testing more parameter-rich priors such as FBD or Coalescent priors.

\subsection{Bayes Factor for Steppe vs. Anatolia}
\begin{table}[!ht]
\centering
% \scriptsize
\begin{tabular}{lccc}
\toprule
%\multirow{2}{*}{Dataset} & \multicolumn{3}{c}{BF} & \multicolumn{3}{c}{Strength} \\\cmidrule(lr){2-4}\cmidrule(lr){5-7}
% & \multicolumn{3}{c}{BF}\\\cmidrule(lr){2-4}
% & FBD & Coalescent & Uniform & FBD & Coalescent & Uniform \\\midrule
Dataset & FBD & Coalescent & Uniform \\\midrule
  \textsc{B1} & 0.138 (Negative)  & ** & 67.043 (Strong)\\% &  & -- & \\
\textsc{B2} & 1.015 (Neutral) & ** & 1022.968 (Very Strong)\\% &  & --  & \\
\textsc{broad} & 88.624 (Strong) &  * & 6728.994 (Very Strong)\\% &  & -- & \\
\textsc{medium} & 18.536 (Positive) & ** & 113.968 (Strong)\\ %&  & -- & \\
\textsc{narrow} & 16.55 (Positive) & * & 27.549 (Strong)\\% &  & -- & \\
\bottomrule
 \end{tabular}
\caption{\label{tab:BF} Bayes Factor Support for the Steppe origin vs. the Anatolian 
origin across different 
datasets and tree priors. * represents a entry where there is no tree in the Prior sample with root age 
that falls within Steppe range. ** indicates those datasets that do not have a posterior and prior root age within 
Steppe range.}
\end{table}

I present the results of the Bayes factor (BF) analysis in table \ref{tab:BF}. In the case of the FBD prior, BF results support the Steppe 
origin hypothesis for all the datasets, except, for the B1 dataset. The corrected datasets clearly support the Steppe hypothesis 
\emph{positively} 
in terms of Bayes Factor in the case of FBD prior. In the case of the uniform prior, all the datasets \emph{support} the Steppe origin 
hypothesis 
over the Anatolian origin hypothesis. In the case of the coalescent 
prior, the Bayes Factor was not possible to calculate since there is no tree in either 
prior or posterior sample that has a root age belonging to the age range of the Steppe hypothesis. Overall, the interpretation of the 
strength of the Bayes Factor 
analysis suggests that appropriate tree priors and corrected datasets \emph{support} the Steppe origin hypothesis of the 
Indo-European language family.

\subsection{Internal node ages}
In this subsection, for each dataset, I compare the inferred dates for the language subgroups 
with the historically attested dates given in table \ref{tab:subgroup}. The uniform tree prior, on an average, overestimates the ages for 
all 
the datasets, except, for the \textsc{narrow} dataset. The predicted ages from the uniform tree prior  come closest to the historical 
ages in the 
case of the \textsc{median} 
dataset. In contrast, \citeauthor{chang2015ancestry} present younger ages for both the \textsc{narrow} (100 years 
on an average) and \textsc{medium} datasets (330 $\pm$ 165 years).

\begin{table}[!ht]
\centering
\scalebox{0.68}{
\begin{tabular}{lcccccc}
\toprule
Subgroup & Historical Age & B1 & B2 & \textsc{Broad} & \textsc{Medium} & \textsc{narrow} \\\midrule
% Subgroup  & Historical Age  & B1  &  & B2  &  & Broad  &  & Medium  &  & Narrow  & \\
Germanic  & 2250 & 2876 [2286-3572] & 2816 [2256-3458] & 2615 [2147-3166] & 2449 [2031-2935] & 2334 [1943-2807]\\
Romance  & 1750 & 2987  [2400-3629] & 2149  [1628-2714] & 1980 [1515-2493] & 1841 [1401-2345] & 1736 [1309-2248]\\
Scandinavian  & 1500 & 1523 [1127-2016] & 1469 [1102-1906] & 1340 [1024-1697] & 1164 [898-1477] & –  \\
Slavic  & 1500 & 1860 [1401-2423] & 1822 [1378-2309] & 1647 [1301-2069] & 1575 [1226-1972] & –  \\
East Baltic  & 1300 & 1584 [914-2356] & 1561 [936-2265] & 1465 [891-2086] & 1460 [892-2115] & –  \\
British Celtic  & 1250 & 1732 [1105-2402] & 1687 [1137-2343] & 1537 [1024-2093] & 1450 [955-2011] & –  \\
Modern  Irish/Scots Gaelic  & 1050 & 1058 [530-1615] & 1052 [589-1620] & 967 [523-1442] & 834 [451-1260] & 829 [442-1290]\\
Persian-Tajik  & 750 & 882 [424-1412] & 842 [386-1360] & 819 [409-1250] & 704 [336-1098] & –  \\\midrule

Average difference & & -394 & -256 & -127 & -15.875 & 50.33
\\\bottomrule

 \end{tabular}
}
\caption{\label{tab:subgroup} The first column shows the name 
of the language subgroup and the second column shows the ages based on historical events. The rest of the columns show the uniform prior's 
median 
ages (in years before present) and 95\% HPD (the numbers in square brackets) age intervals across the five datasets. The last row shows the 
average 
difference between historical ages and predicted ages. The historical ages are 
obtained from \citet[226]{chang2015ancestry}. The 
East Baltic group consists of Lithuanian and Latvian. The British Celtic group consists of Cornish, 
Breton, and Welsh. The Romance group consists of all the Romance languages except Latin.}
\end{table}

\begin{figure}[!ht]
\centering
% \scalebox{0.25}{
 \includegraphics[height=1.15\linewidth,width=\linewidth]{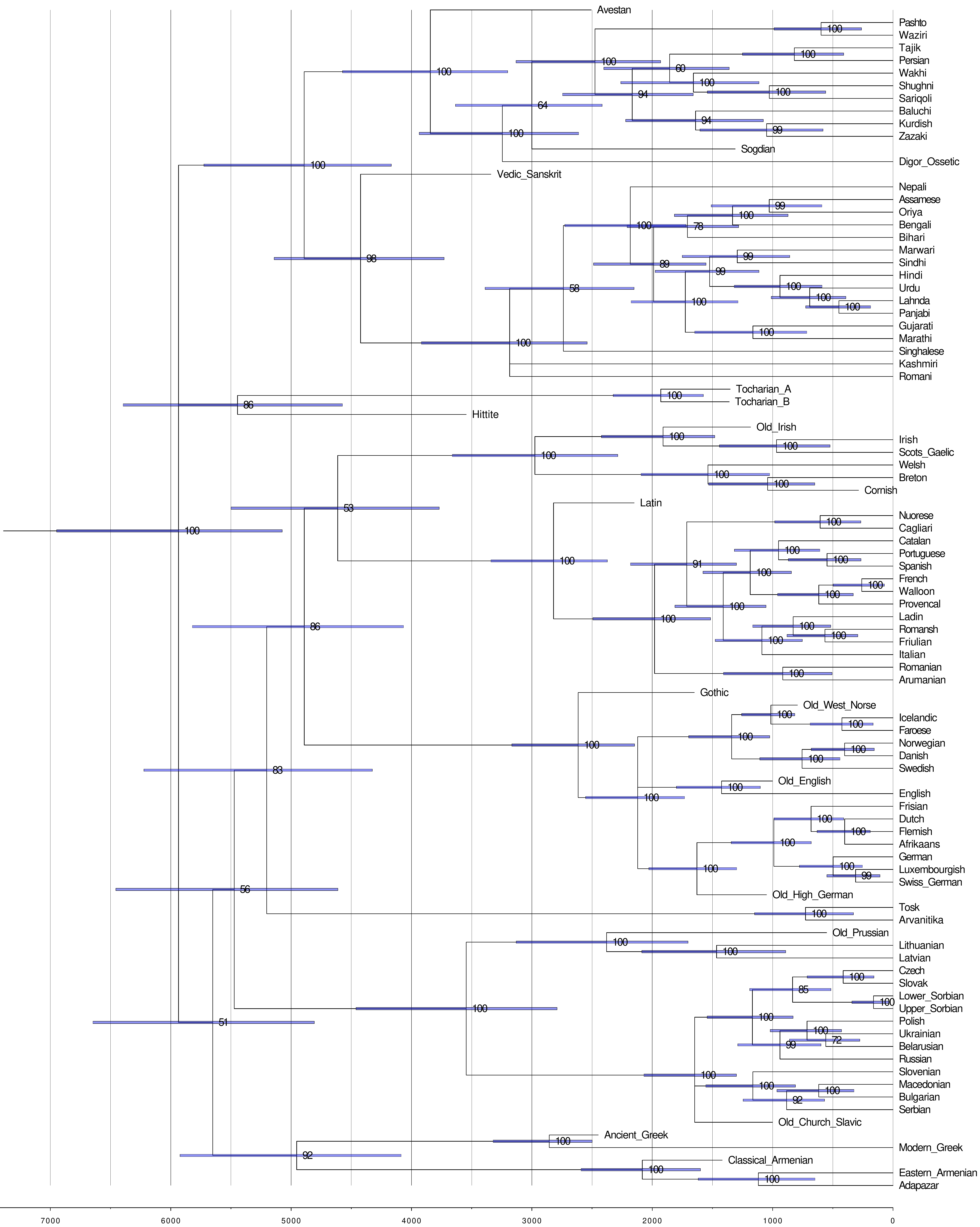}
\caption{The majority-rule consensus tree inferred using uniform prior for the \textsc{broad} dataset. The numbers at 
each internal node shows the 
support for the subtree in the posterior sample. The blue bars show the 95\% 
HPD intervals for the node ages. The time scale shows the height of the tree in 
terms of age.}
\label{fig:unifBroad}
\end{figure}

\subsection{Relevance of clade constraints}
Both \citet{bouckaert2012mapping} and \citet{chang2015ancestry} constrain the 
topologies in tree search through clade constraints. For 
instance, a Germanic clade constraint would 
mean that the Bayesian software would only sample those trees that place all 
the Germanic languages under a single node. Both the studies do not follow the same set of topological constraints when inferring the dates 
of Indo-European language family. \citet{chang2015ancestry} apply a stricter set of constraints---derived from the linguistic knowledge 
of Indo-European language family---than those of \citet{bouckaert2012mapping}.
%In effect, constraining the complete topology 
%implies that the Bayesian software only attempts to infer the age of the 
%Indo-European family. 
In this paper, I do not employ 
any clade constraints and allow the software to automatically infer the tree topology from the datasets.\footnote{I note that the clade 
constraint information is 
derived from historical linguistics research that is limited to language families such as Indo-European, Dravidian, Uralic, 
Austronesian, and Sino-Tibetan with long tradition of classical comparative linguistic research \citep{campbell2008language}.}

I present the majority rule consensus tree inferred using uniform prior 
for the \textsc{broad} dataset in 
figure \ref{fig:unifBroad}.\footnote{All the trees presented in this paper are visualized using FigTree \citep{rambaut2007figtree}.} The 
majority rule consensus tree retrieves the well-established language subgroups such as Balto-Slavic, Greek, 
Indo-Iranian, Germanic, and Italo-Celtic 
correctly. I observe that all the consensus trees (appendix) retreive the subgroups correctly without being supplied as constraints to 
the phylogenetic software.

\paragraph{Position of Anatolian and Tocharian languages}
There is a general consensus among the Indo-European scholars that the Anatolian language group was the first branch to split from the 
Proto-Indo-European stage, after which, the Tocharian language group was the second to split off from the post-Anatolian Indo-European 
languages \citep{ringe2002indo}. In fact, 
\citeauthor{chang2015ancestry} supply this linguistic knowledge as two constraints to the Bayesian software: Nuclear Indo-European group 
consisting of all the non-Anatolian languages; and, Inner 
Indo-European group consisting 
of all the Nuclear Indo-European languages excluding Tocharian languages. I observe that the majority consensus trees constructed from the 
analyses inferred with uniform tree prior always groups both 
the Anatolian and Tocharian languages as distinct subgroups unified under the same internal node which is directly connected to the root 
node. This is also true in the case of the majority consensus 
tree inferred when the coalescent tree prior is applied to the B2 dataset. The majority consensus trees constructed from FBD tree prior's 
analyses always show that the Anatolian languages 
were the first to split off, followed by the branching of the Tocharian languages from the post-Anatolian Indo-European complex. This 
observation also holds for the the majority consensus trees 
inferred with colaescent tree priors applied to B1, \textsc{broad, medium}, and \textsc{narrow} datasets.\\

\noindent In conclusion, the majority consensus trees suggest that the well-established Indo-European subgroups can be 
inferred directly, and need not be supplied beforehand. The exact placement of the well-established subgroups with respect to each other within the 
Inner 
Indo-European clade is a topic of research among scholars and has to be determined to full satisfaction \citep{anthony2015indo}.

% The conclusion from this experiment 
% is that the Bayesian software does not require the clade constraints to infer the right 
% Indo-European tree.

\subsection{Relevance of ancestry constraints}
\citet{chang2015ancestry} introduced ancestry constraints into their phylogenetic analysis, which, then, supported the Steppe origin 
hypothesis. The application of the FBD prior can be used to verify if the ancestry constraints can be inferred from the data. The FBD 
prior can infer whether an ancient language is an ancestral language or a tip in the tree. However, the majority rule consensus trees 
inferred from all the datasets using FBD tree prior do not show any support for the ancestry relationships enforced as constraints by 
\citet{chang2015ancestry}. I examined the log files of 
the MCMC runs and found that the MCMC proposal move (\texttt{delete-branch}) in MrBayes supporting the placement of an ancient language as 
an
internal node was never accepted during the MCMC sampling. At least, based on trees inferred from lexical datasets, I conclude that the FBD 
prior does not infer any ancestry relations employed by \citet{chang2015ancestry}.

\section{Conclusion}\label{sec:concl}
In this paper, I addressed the question of the effect of tree priors in Bayesian phylogenetic analysis and found the following.
\begin{itemize}
 
\item  The model comparison results suggest 
that both Uniform and FBD priors show better fit to the datasets of the Indo-European language family than the coalescent prior. Therefore, based on the Bayes 
Factor analysis, I conclude that the Steppe hypothesis is supported by FBD and Uniform priors for majority of the datasets.
\item The FBD tree prior does not infer any ancestry relation from any of the datasets suggesting that the 
lexical datasets used in the paper does not have signal for ancestry relations.
\item I also observe that the Bayesian inference program can infer well-established subgroups correctly from the data and need not be supplied 
beforehand.
\item Finally, the 
experiments reported in the paper suggest that right tree priors and corrected cognacy judgments are important for estimating the phylogeny and the 
age of Indo-European language family.
\end{itemize}

\section*{Acknowledgments}
% \small
The paper would not have been possible without the 
continuous 
support of Igor Yanovich, Søren Wichmann, Chris Bentz, Gerhard Jäger, Johann-Mattis List, Richard Johansson, Lilja Øvrelid, Sowmya 
Vajjala, Çağrı Çöltekin, and Aparna 
Subhakari. I thank Remco Bouckaert, Johannes Wahle, Armin Buch, Johannes Dellert, Marisa Köllner, Roland Mühlenbernd, and Vijayaditya Peddinti for all the 
comments and discussions that improved the paper. Finally, 
I thank the anonymous reviewers for all the comments which helped 
improved the paper. One of the reviewers provided extensive 
comments regarding the models and results which helped improve the paper. All the remaining errors are mine. The author is 
supported by BIGMED and ERC 
Advanced Grant 324246 EVOLAEMP, which is gratefully acknowledged.

% धर्मो रक्षति रक्षितः

\clearpage
% \newpage
\bibliographystyle{BrillLDC}
\bibliography{myreflnks}

% \clearpage
 \newpage
\appendix
 \section{Coalescent Prior}
\begin{figure}[!ht]
\centering
\includegraphics[height=1.25\linewidth,width=\linewidth]{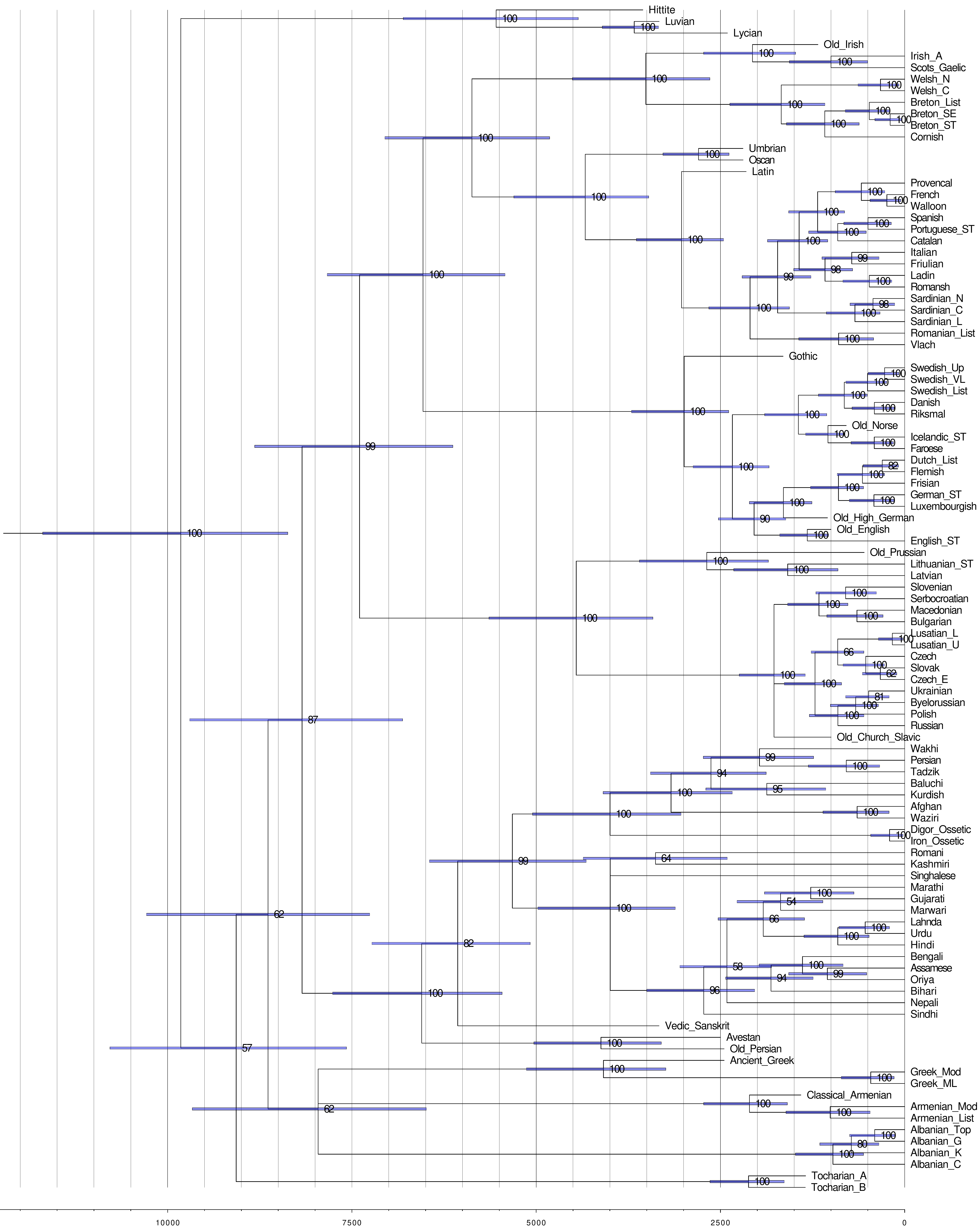}
\caption{The majority-rule consensus tree for B1 dataset. The numbers at each 
internal node shows the support for the subtree in the posterior sample. The 
blue bars show the 95\% HPD intervals for the node ages. The time scale shows 
the height of the tree in terms of age.}
\label{fig:coalB1}
\end{figure}

\clearpage

\begin{figure}[!ht]
\centering
\includegraphics[height=\paperwidth,width=\linewidth]{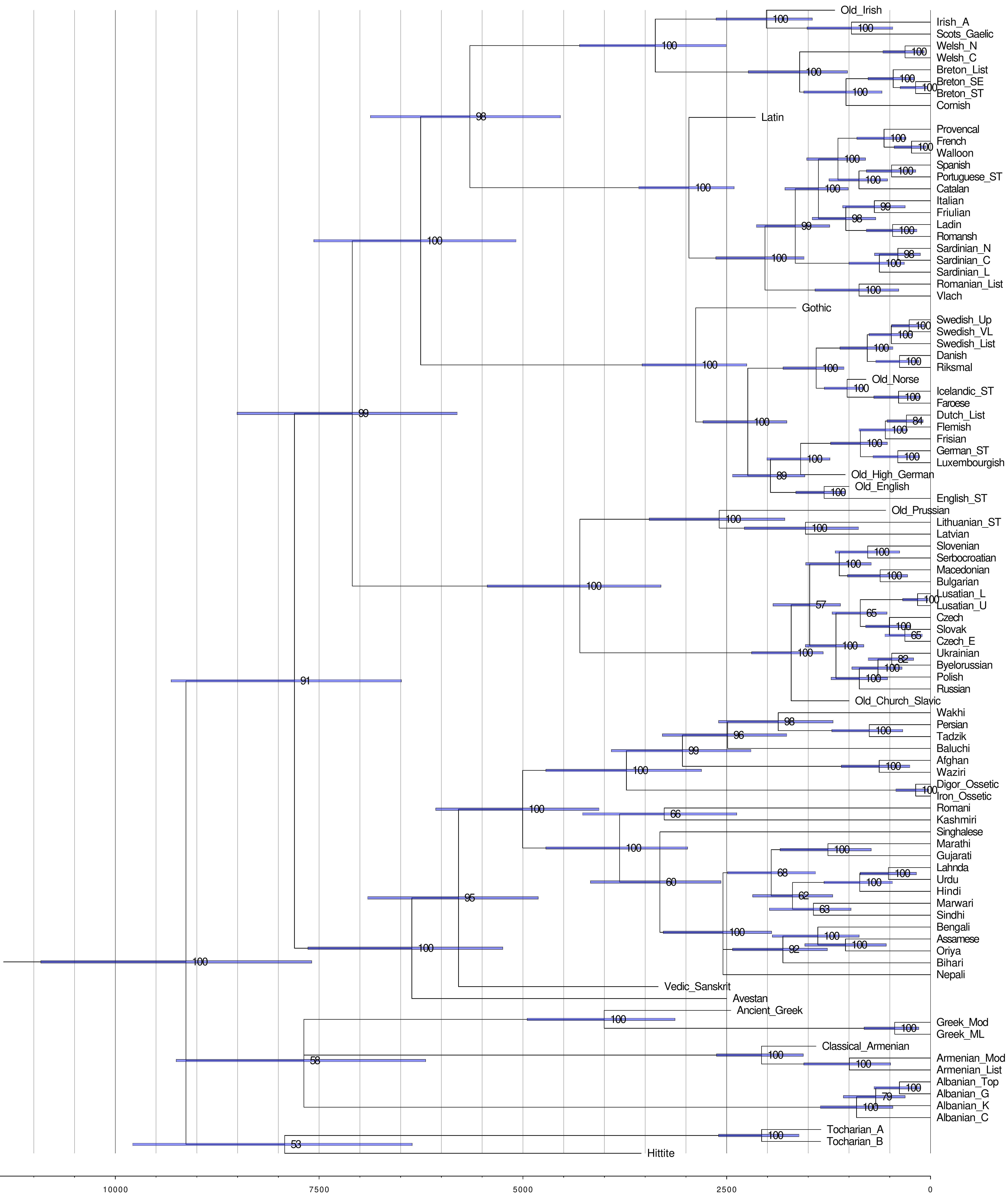}
\caption{The majority-rule consensus tree for B2 dataset. The numbers at each 
internal node shows the support for the subtree in the posterior sample. The 
blue bars show the 95\% HPD intervals for the node ages. The time scale shows 
the height of the tree in terms of age.}
\label{fig:coalB2}
\end{figure}

\clearpage

\begin{figure}[!ht]
\centering
\includegraphics[height=\paperwidth,width=\linewidth]{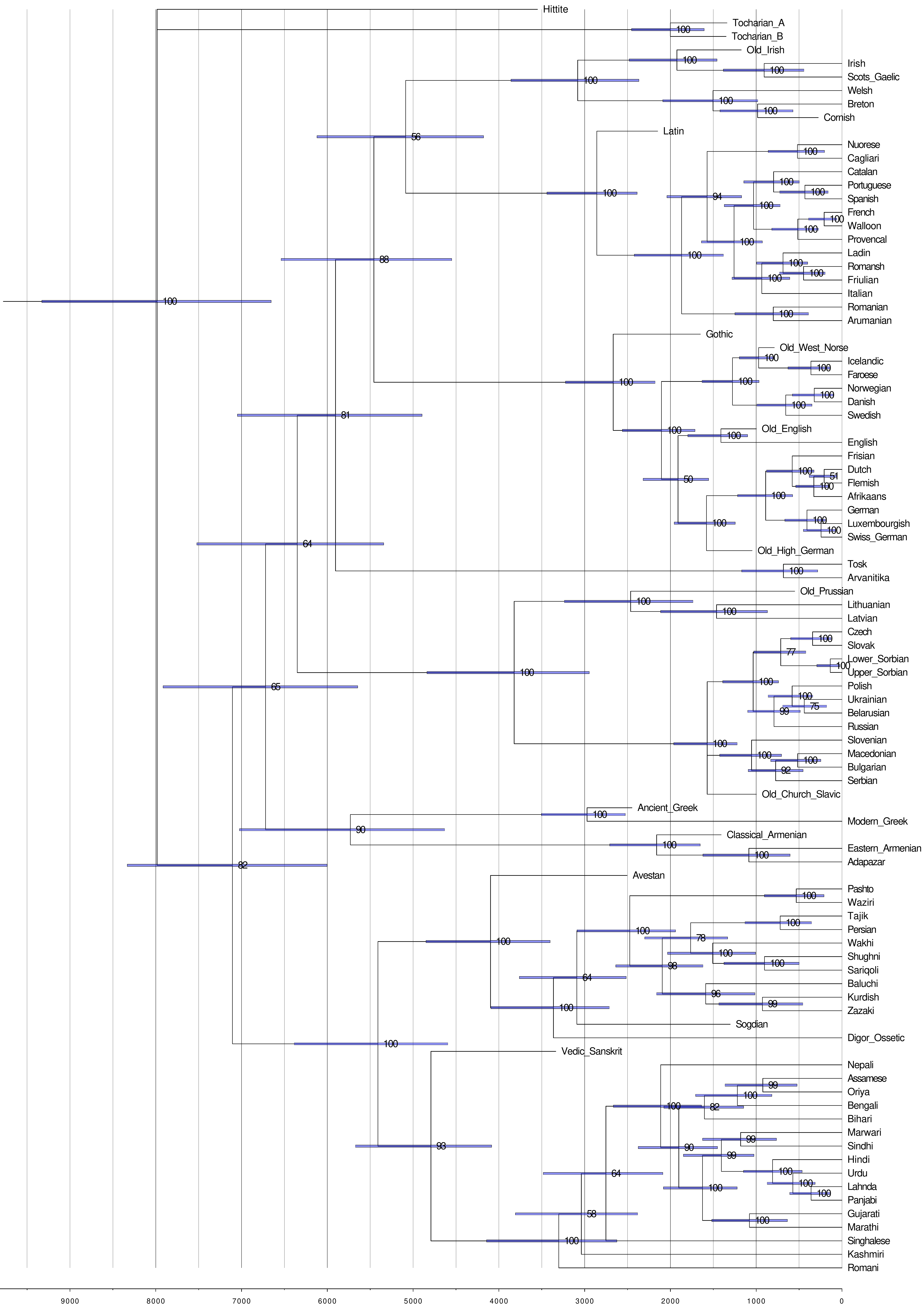}
\caption{The majority-rule consensus tree for \textsc{broad} dataset. The numbers at each 
internal node shows the support for the subtree in the posterior sample. The 
blue bars show the 95\% HPD intervals for the node ages. The time scale shows 
the height of the tree in terms of age.}
\label{fig:coalBroad}
\end{figure}

\clearpage

\begin{figure}[!ht]
\centering
\includegraphics[height=\paperwidth,width=\linewidth]{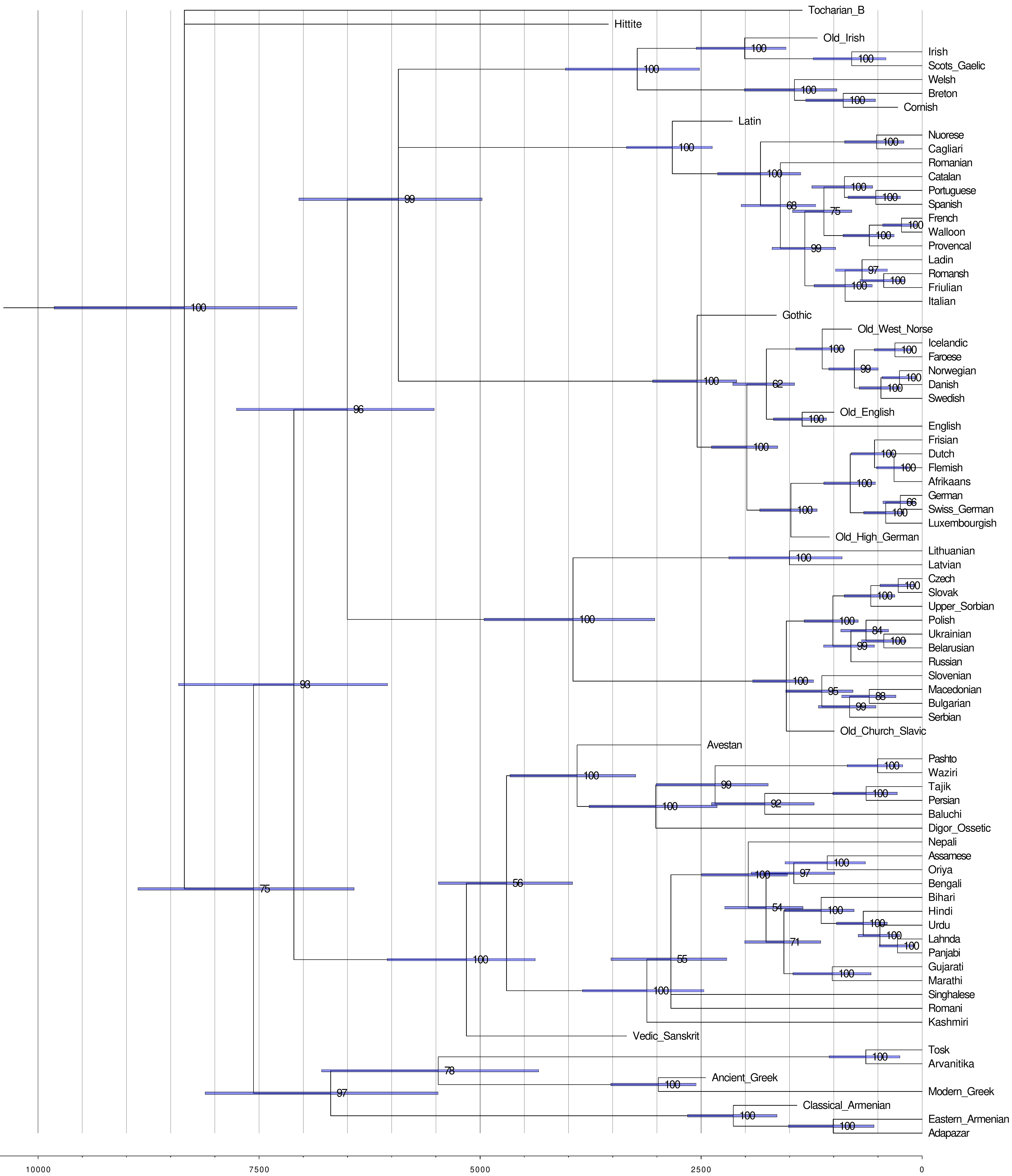}
\caption{The majority-rule consensus tree for \textsc{medium} dataset. The numbers at each 
internal node shows the support for the subtree in the posterior sample. The 
blue bars show the 95\% HPD intervals for the node ages. The time scale shows 
the height of the tree in terms of age.}
\label{fig:coalMedium}
\end{figure}

\clearpage

\begin{figure}[!ht]
\centering
\includegraphics[height=\linewidth,width=\linewidth]{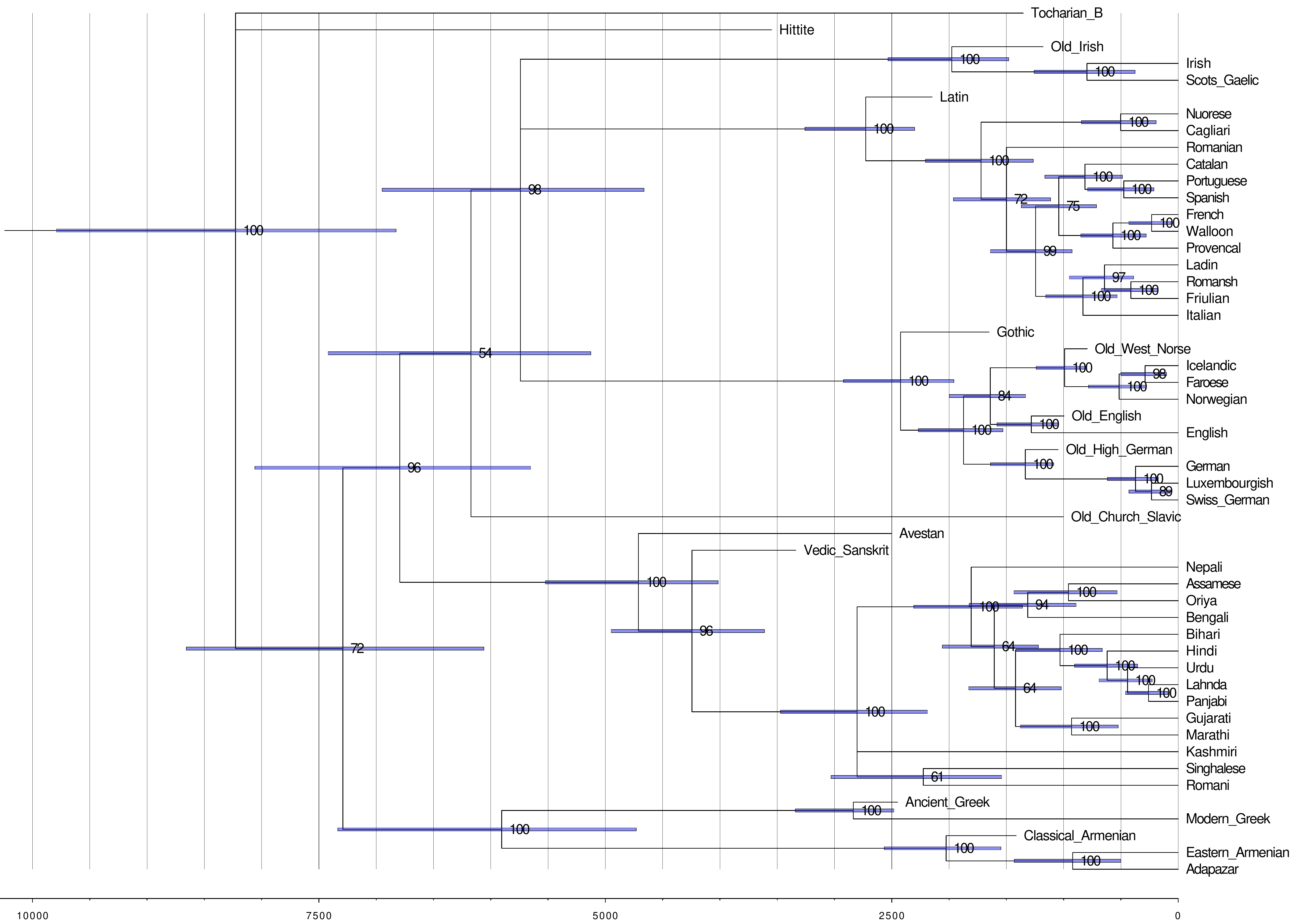}
\caption{The majority-rule consensus tree for \textsc{narrow} dataset. The numbers at each 
internal node shows the support for the subtree in the posterior sample. The 
blue bars show the 95\% HPD intervals for the node ages. The time scale shows 
the height of the tree in terms of age.}
\label{fig:coalNarrow}
\end{figure}

\newpage

 \section{FBD Prior}
 \begin{figure}[!ht]
\centering
\includegraphics[height=1.25\linewidth,width=\linewidth]{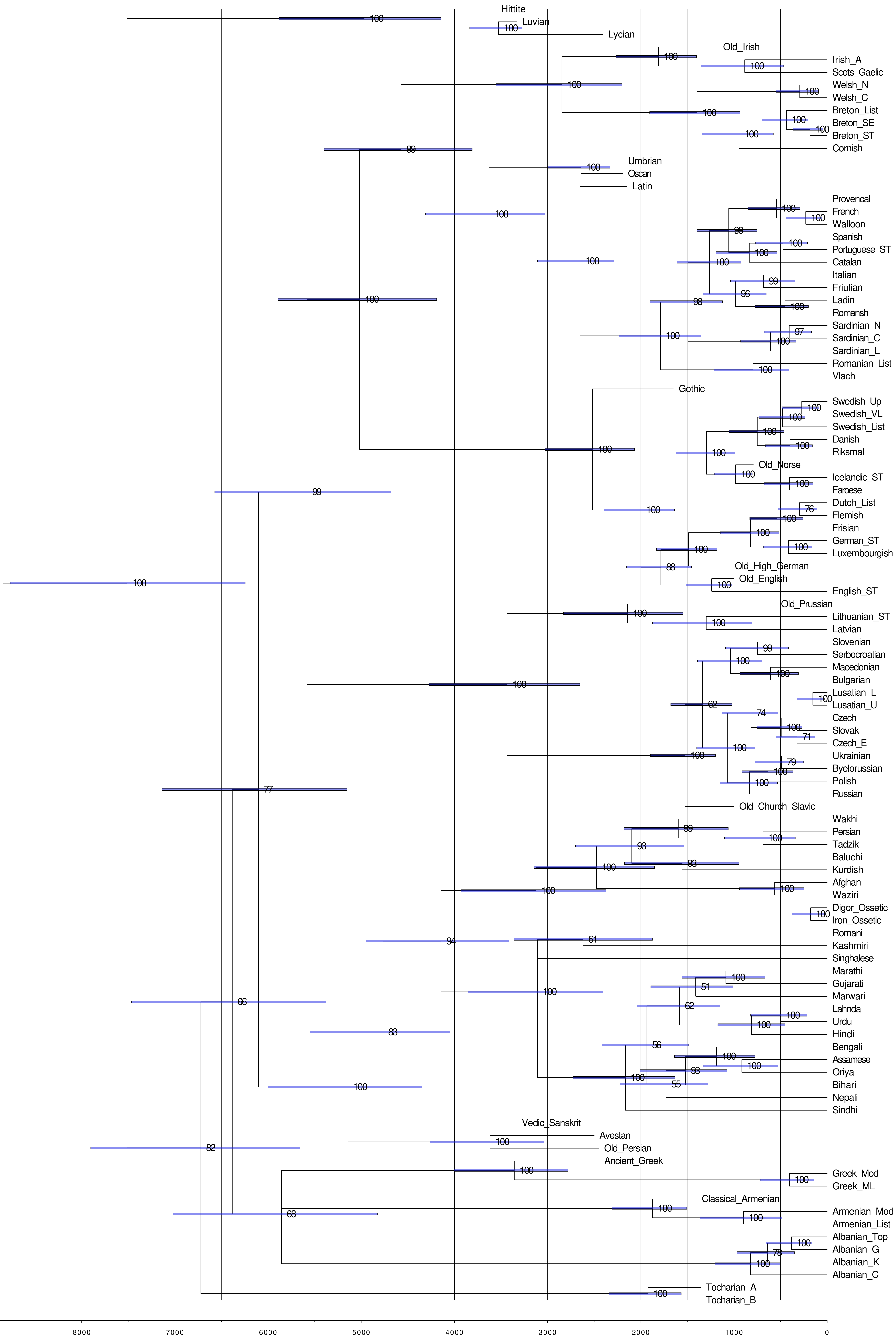}
\caption{The majority-rule consensus tree for B1 dataset. The numbers at each 
internal node shows the support for the subtree in the posterior sample. The 
blue bars show the 95\% HPD intervals for the node ages. The time scale shows 
the height of the tree in terms of age.}
\label{fig:fbdB1}
\end{figure}

\clearpage

\begin{figure}[!ht]
\centering
\includegraphics[height=\paperwidth,width=\linewidth]{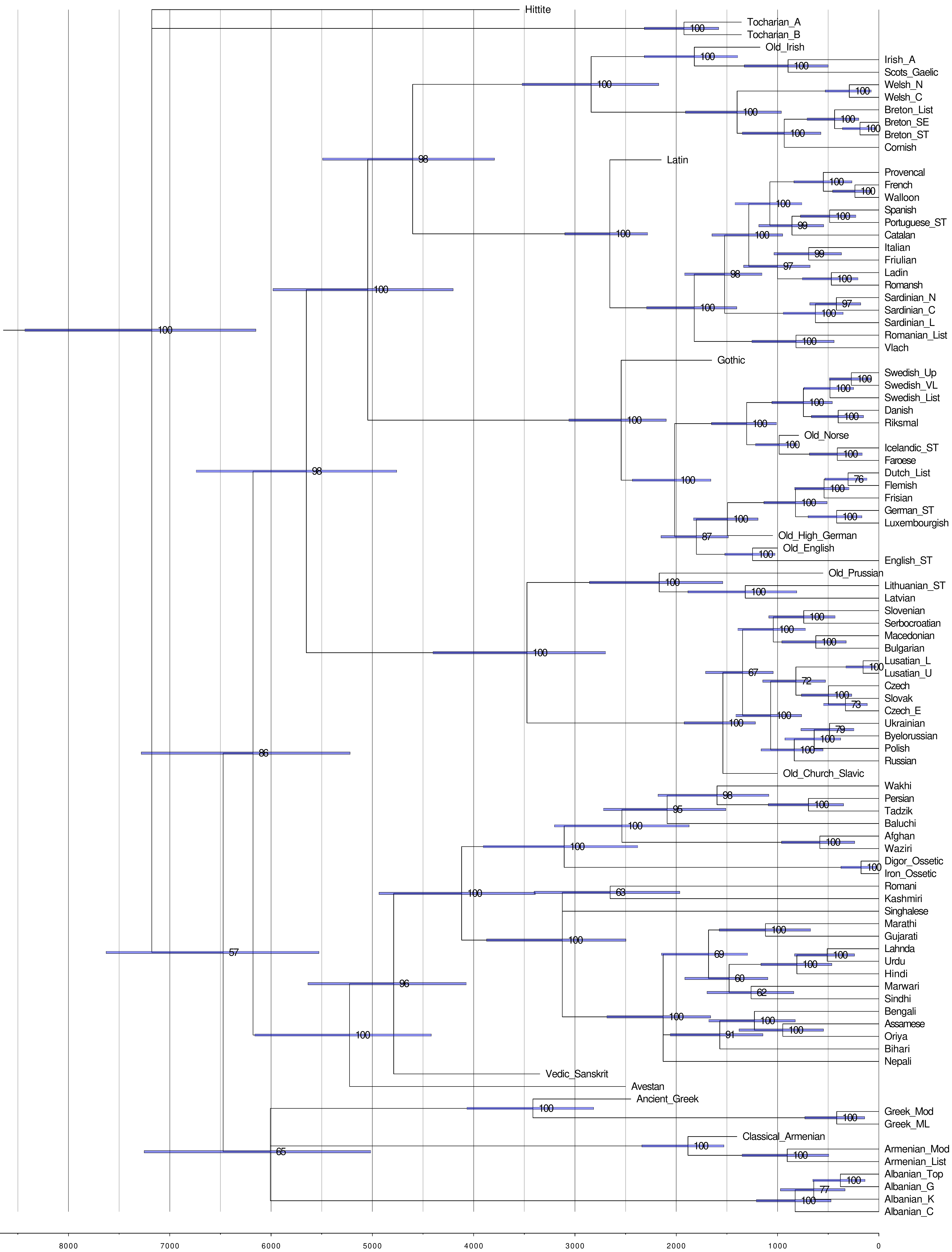}
\caption{The majority-rule consensus tree for B2 dataset. The numbers at each 
internal node shows the support for the subtree in the posterior sample. The 
blue bars show the 95\% HPD intervals for the node ages. The time scale shows 
the height of the tree in terms of age.}
\label{fig:fbdB2}
\end{figure}

\clearpage

\begin{figure}[!ht]
\centering
\includegraphics[height=\paperwidth,width=\linewidth]{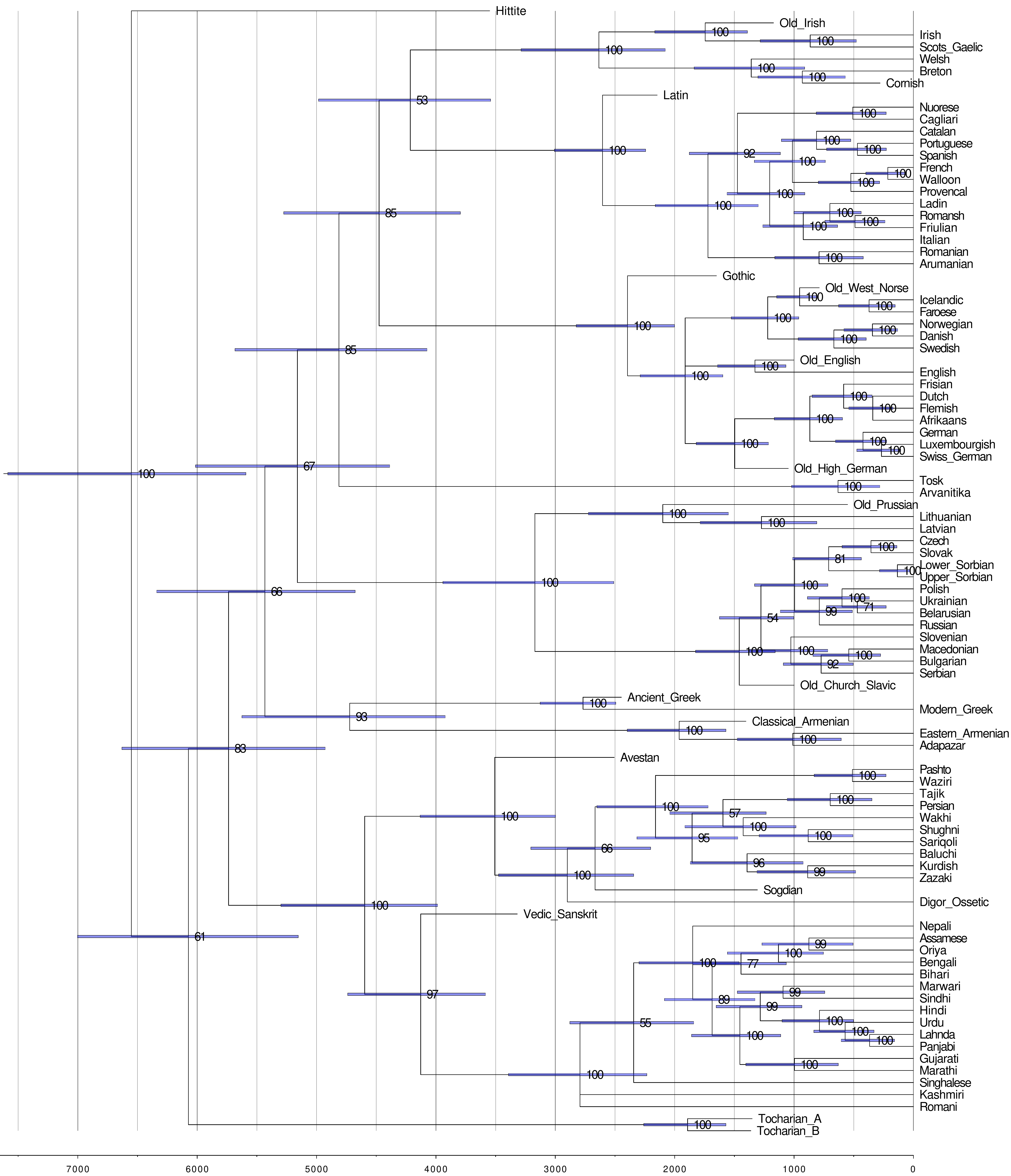}
\caption{The majority-rule consensus tree for \textsc{broad} dataset. The numbers at each 
internal node shows the support for the subtree in the posterior sample. The 
blue bars show the 95\% HPD intervals for the node ages. The time scale shows 
the height of the tree in terms of age.}
\label{fig:fbdBroad}
\end{figure}

\clearpage

\begin{figure}[!ht]
\centering
\includegraphics[height=\paperwidth,width=\linewidth]{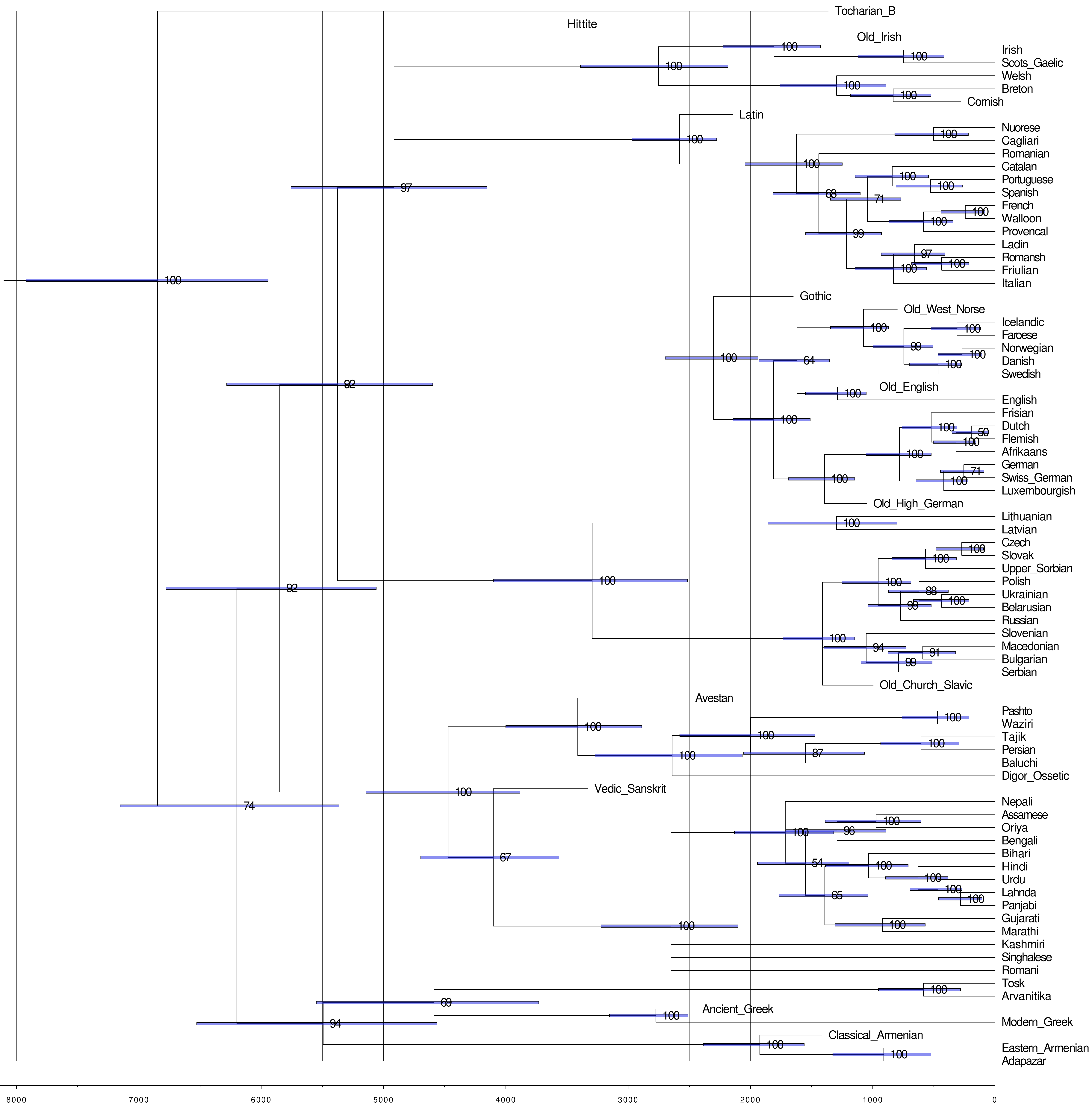}
\caption{The majority-rule consensus tree for \textsc{medium} dataset. The numbers at each 
internal node shows the support for the subtree in the posterior sample. The 
blue bars show the 95\% HPD intervals for the node ages. The time scale shows 
the height of the tree in terms of age.}
\label{fig:fbdMedium}
\end{figure}

\clearpage

\begin{figure}[!ht]
\centering
\includegraphics[height=\paperwidth,width=\linewidth]{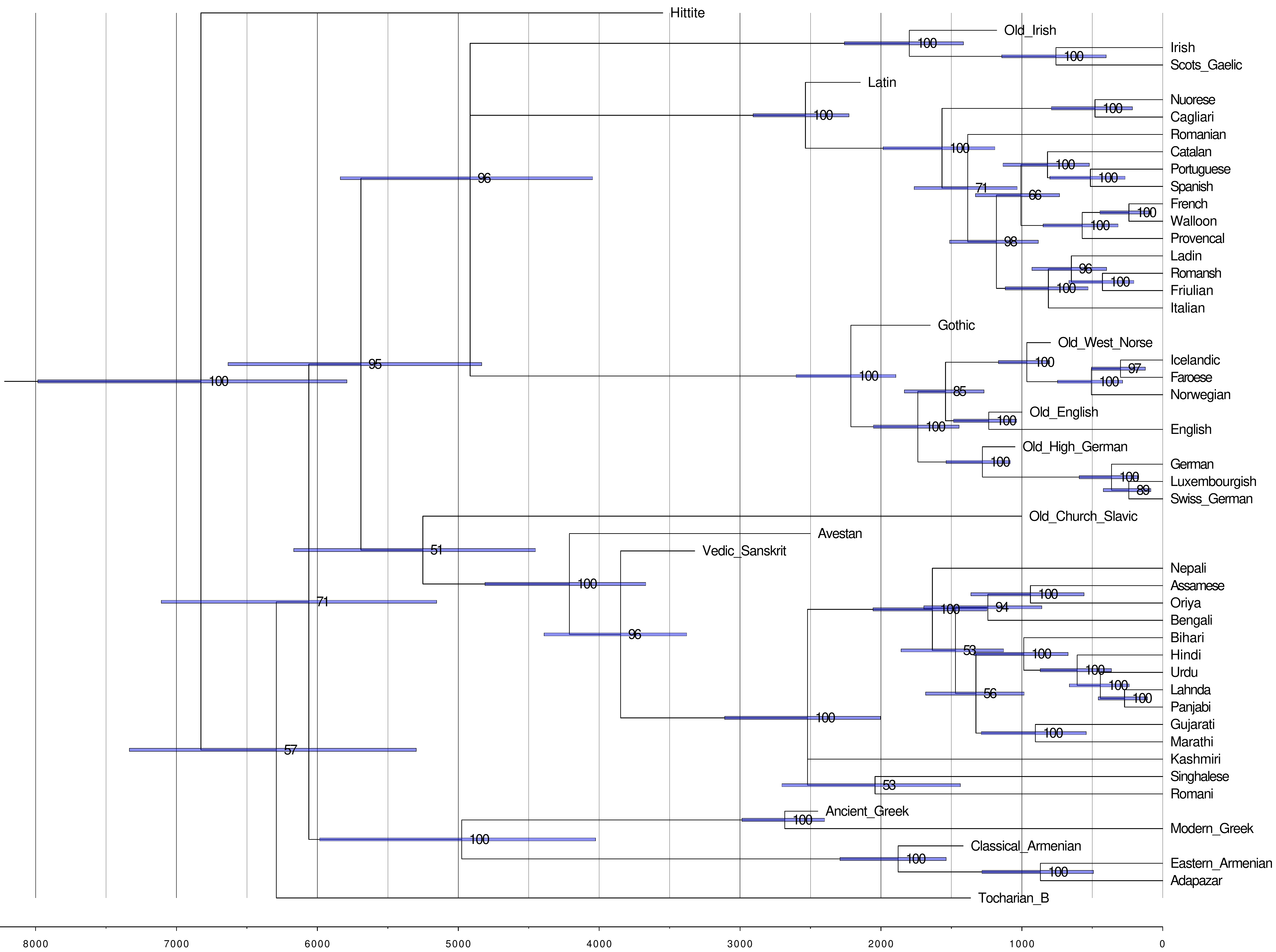}
\caption{The majority-rule consensus tree for \textsc{narrow} dataset. The numbers at each 
internal node shows the support for the subtree in the posterior sample. The 
blue bars show the 95\% HPD intervals for the node ages. The time scale shows 
the height of the tree in terms of age.}
\label{fig:fbdNarrow}
\end{figure}
 
 \clearpage
\newpage
\section{Uniform Prior}
 \begin{figure}[!ht]
\centering
\includegraphics[height=1.25\linewidth,width=\linewidth]{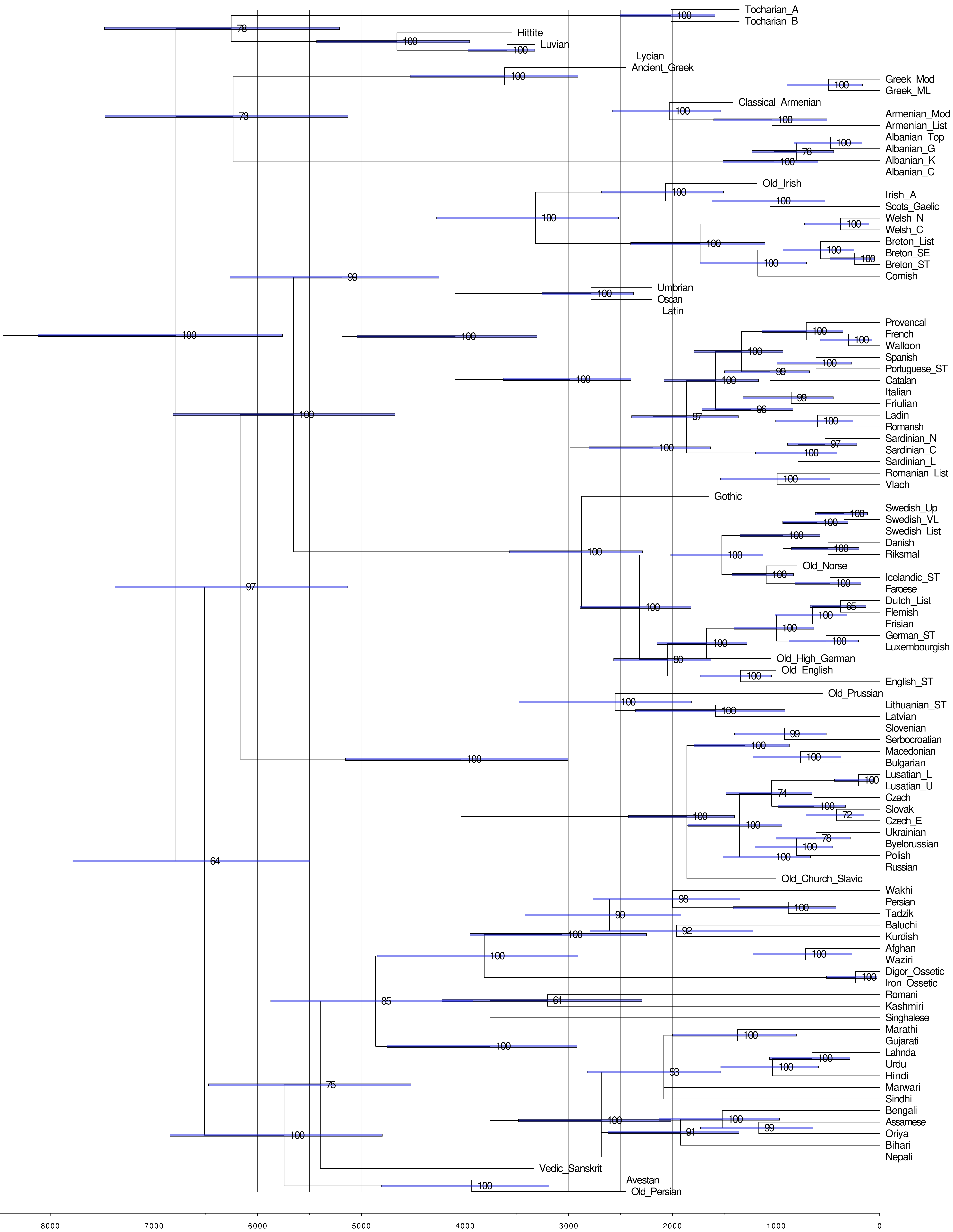}
\caption{The majority-rule consensus tree for B1 dataset. The numbers at each 
internal node shows the support for the subtree in the posterior sample. The 
blue bars show the 95\% HPD intervals for the node ages. The time scale shows 
the height of the tree in terms of age.}
\label{fig:unifB1}
\end{figure}

\clearpage

\begin{figure}[!ht]
\centering
\includegraphics[height=\paperwidth,width=\linewidth]{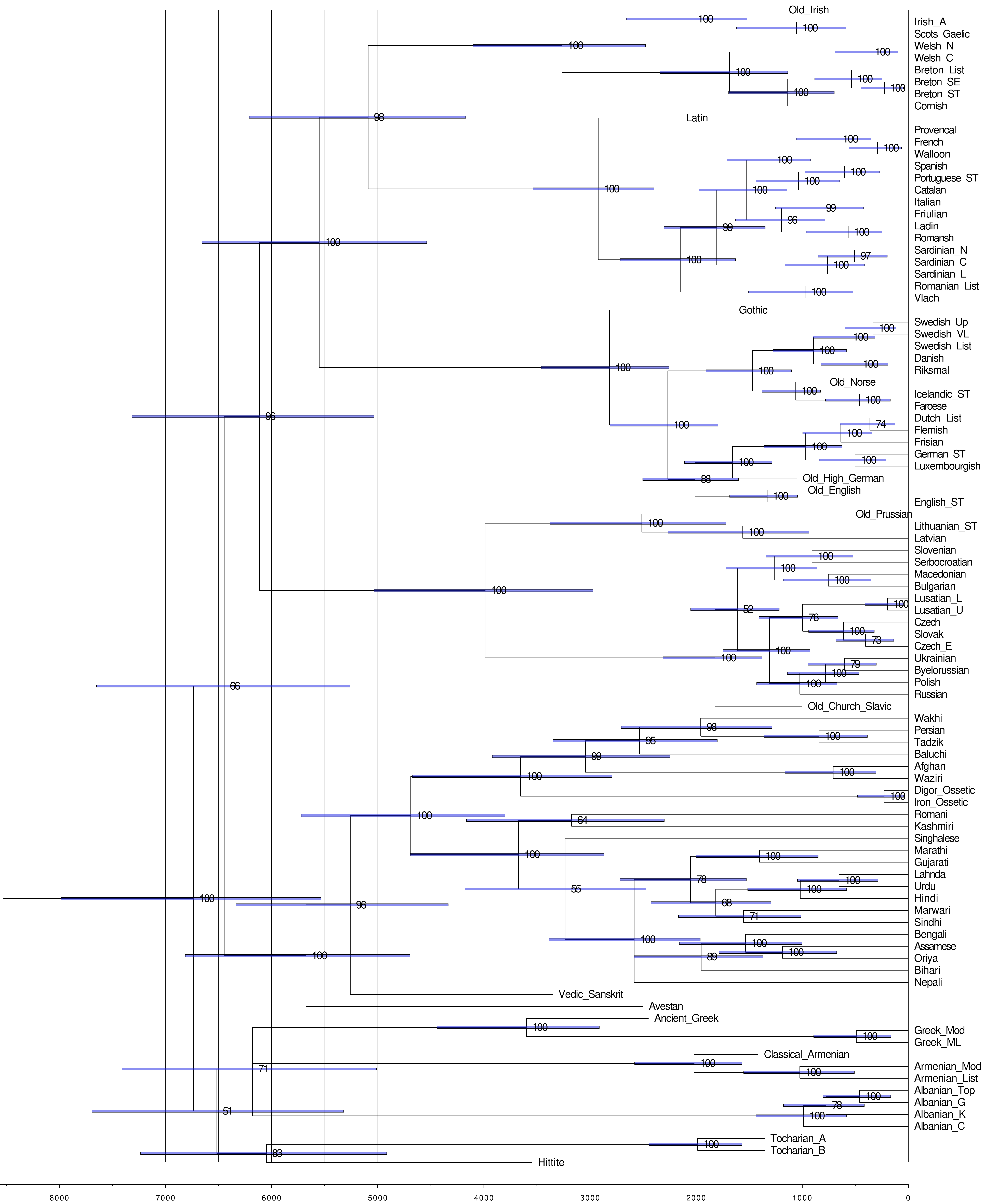}
\caption{The majority-rule consensus tree for B2 dataset. The numbers at each 
internal node shows the support for the subtree in the posterior sample. The 
blue bars show the 95\% HPD intervals for the node ages. The time scale shows 
the height of the tree in terms of age.}
\label{fig:unifB2}
\end{figure}

\clearpage

\begin{figure}[!ht]
\centering
\includegraphics[height=\paperwidth,width=\linewidth]{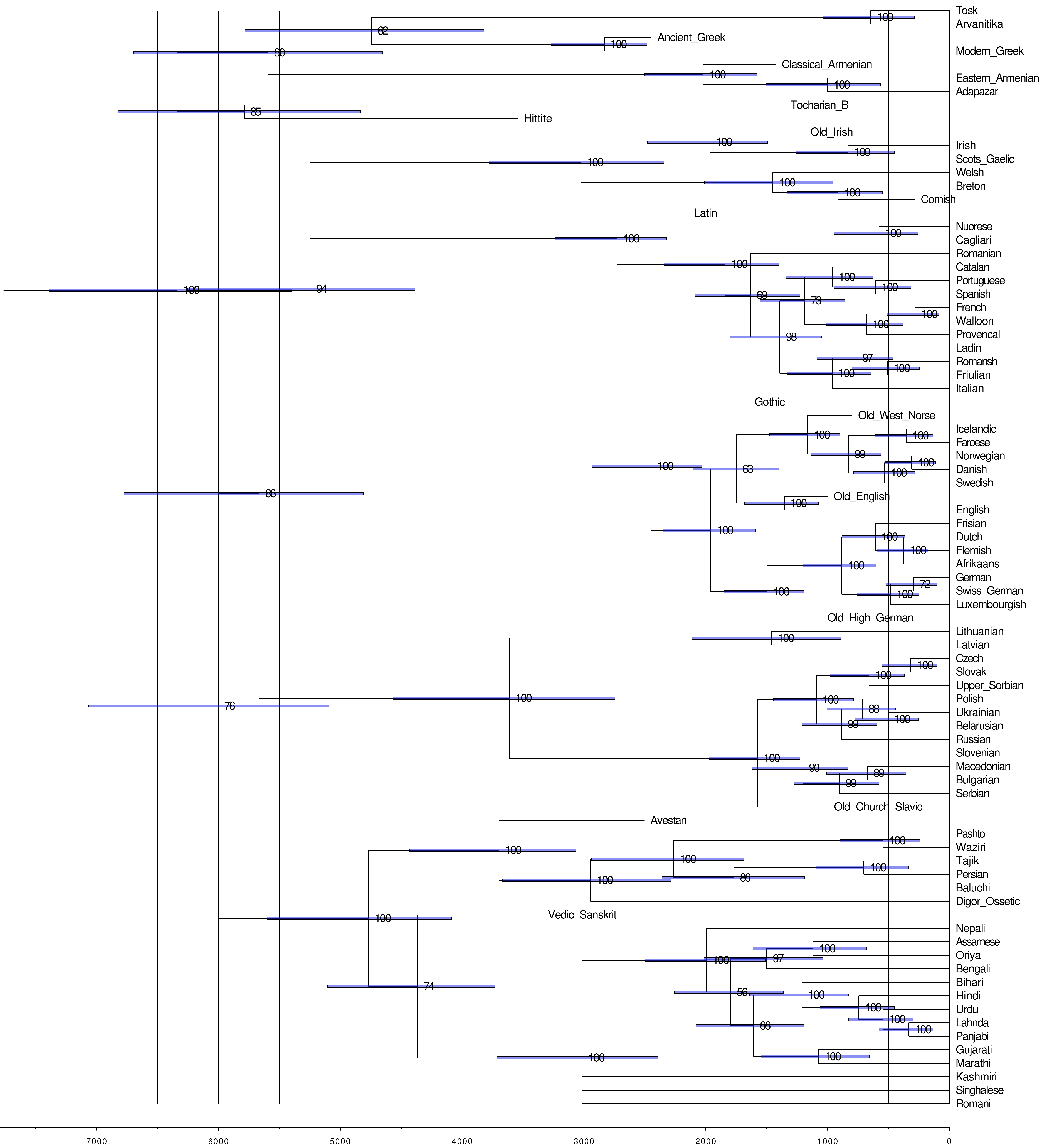}
\caption{The majority-rule consensus tree for \textsc{medium} dataset. The numbers at each 
internal node shows the support for the subtree in the posterior sample. The 
blue bars show the 95\% HPD intervals for the node ages. The time scale shows 
the height of the tree in terms of age.}
\label{fig:unifMedium}
\end{figure}

\clearpage

\begin{figure}[!ht]
\centering
\includegraphics[height=\linewidth,width=\linewidth]{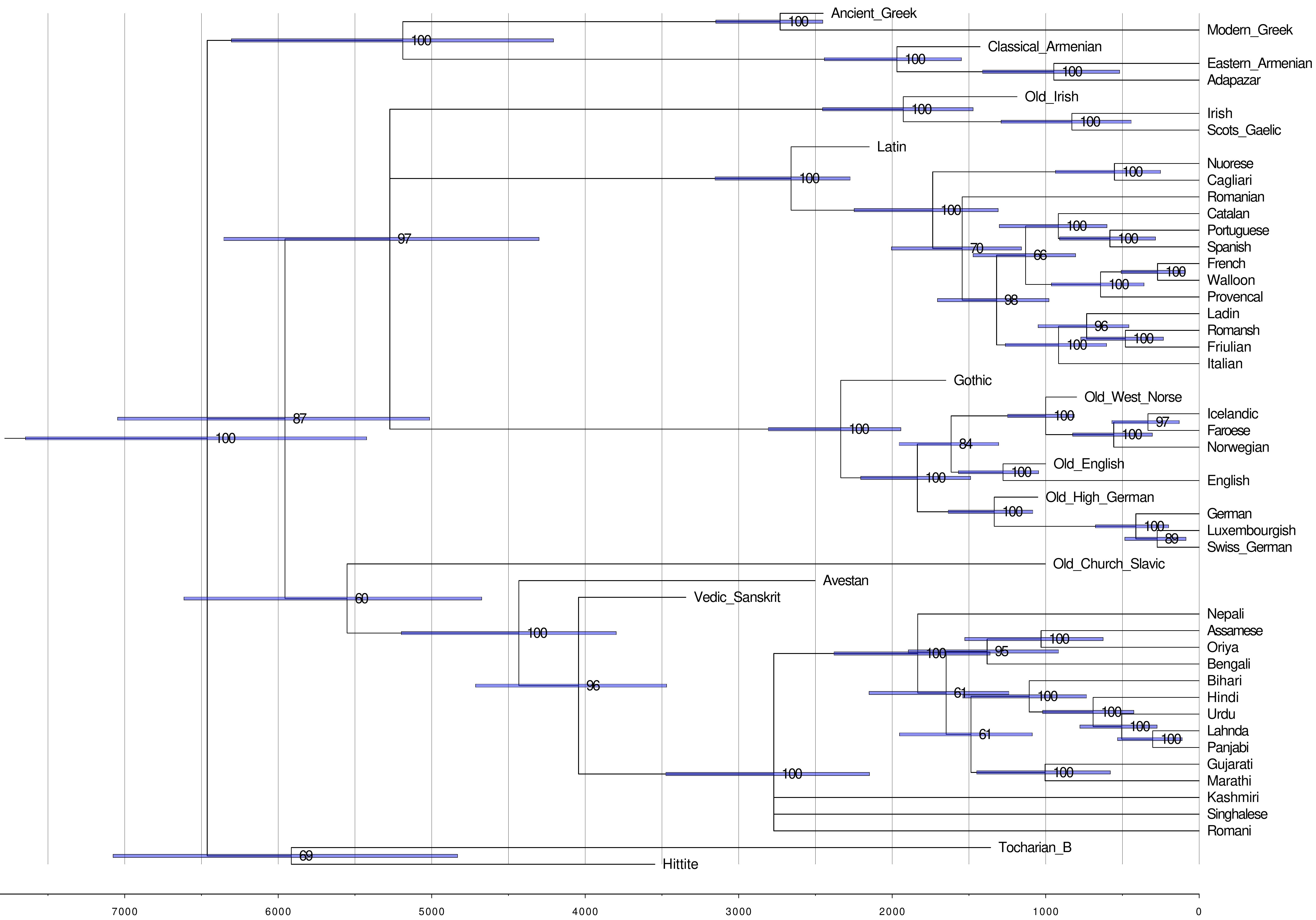}
\caption{The majority-rule consensus tree for \textsc{narrow} dataset. The numbers at each 
internal node shows the support for the subtree in the posterior sample. The 
blue bars show the 95\% HPD intervals for the node ages. The time scale shows 
the height of the tree in terms of age.}
\label{fig:unifNarrow}
\end{figure}

\end{document}